\documentclass[11pt,twoside]{article}

\usepackage{epsfig,amsfonts,color}
\usepackage{amsmath}

\usepackage{amssymb, palatino, geometry,url}
\usepackage{algorithmic}
\usepackage[noresetcount,lined,boxed]{algorithm2e} 
\newcommand{\mynote}[1]{{ \color{black}{#1}}}
\usepackage[colorlinks=true,linkcolor=blue,citecolor=blue,urlcolor=blue]{hyperref}

\usepackage{fancyhdr}
\newcommand{\be}{\begin{equation}}
\newcommand{\ee}{\end{equation}}
\newcommand{\bea}{\begin{eqnarray}}
\newcommand{\eea}{\end{eqnarray}}

\newcommand{\bvec}{\left(\begin{array}{c}}
\newcommand{\evec}{\end{array}\right)}
\newcommand{\bsub}{\begin{subequations}}
\newcommand{\esub}{\end{subequations}}

\usepackage{lineno}

\usepackage{amsthm}

\usepackage{appendix}
\usepackage{multirow}

\usepackage{caption}
\usepackage{subcaption}

\begin{document}

\title{Convolutional Neural Networks: \\ Basic Concepts and Applications in Manufacturing}

\author{Shengli Jiang${}^{\dag}$, Shiyi Qin${}^{\dag}$, Joshua L. Pulsipher${}^{\ddag}$, and Victor M. Zavala${}^{\dag}$\thanks{Corresponding Author: victor.zavala@wisc.edu}\\
  {\small ${}^{\dag}$Department of Chemical and Biological Engineering}\\
 {\small \;University of Wisconsin-Madison, 1415 Engineering Dr, Madison, WI 53706, USA}\\ 
   {\small ${}^{\ddag}$Department of Chemical Engineering}\\
 {\small \;Carnegie Mellon University, 5000 Forbes Ave, Pittsburgh, PA  15213, USA} }
 \date{}
\maketitle

\vspace{-0.2in}

\begin{abstract}
We discuss basic concepts of convolutional neural networks (CNNs) and outline uses in manufacturing. We begin by discussing how different types of data objects commonly encountered in manufacturing (e.g., time series, images, micrographs, videos, spectra, molecular structures) can be represented in a flexible manner using tensors and graphs. We then discuss how CNNs use convolution operations to extract informative features (e.g., geometric patterns and textures) from the such representations to predict emergent properties and phenomena and/or to identify anomalies. We also discuss how CNNs can exploit color as a key source of information, which enables the use of modern computer vision hardware (e.g., infrared, thermal, and hyperspectral cameras).  We illustrate the concepts using diverse case studies arising in spectral analysis, molecule design, sensor design, image-based control, and multivariate process monitoring.
\end{abstract}


\section{Introduction}

Manufacturing is seeing an increasing use of real-time sensing and instrumentation technologies that generate data in the form of images/video (e.g., infrared and thermal),  vibration/audio, and other complex data forms such as chemical spectra and geometrical structures (e.g., 3D printed objects, synthesized molecules, crystals). Manufacturing is also seeing the increasing use of automation systems that aim to exploit such data to make decisions (e.g., optimize production and detect anomalies). Moreover, modern automation systems are being designed to take instructions/targets in the form of complex data objects  (e.g., voice, text, chemical spectra, and molecular structures). 
\\

Modern automation systems used in manufacturing embed highly sophisticated computing workflows that use tools from data science and machine learning (ML) to extract and interpret actionable information from complex data streams. Such workflows resemble those used in other advanced technologies such as autonomous vehicles (e.g., aerial, terrestrial, and aquatic) and robotics.  Moreover, such workflows begin to resemble human systems in which visual, auditive, tactile, and olfactory signals (data) are routinely used to make decisions. For instance, the human olfactory system generates signals when exposed to specific chemical structures and such signals are processed and interpreted by the brain to detect anomalies. Similarly, the human visual system generates interpretable signals when exposed to objects with specific geometrical and color features and our auditory system generates interpretable signals when exposed with specific frequencies.  As such, from a conceptual stand-point, we can see that sensing and data science technologies are enabling increasing convergence between industrial (artificial) and human (natural) perception and decision-making.  This opens new and exciting opportunities to synergize human and artificial intelligence with the ultimate goal of making manufacturing more efficient, safe, sustainable, and reliable.  
\\

In this chapter, we focus on ML technologies that enable information extraction from complex data sources commonly encountered in manufacturing. Specifically, we review basic concepts of convolutional neural networks (CNNs) and outline how these tools can be used to conduct diverse decision-making tasks of interest in manufacturing.  At their core, CNNs use a powerful and flexible mathematical operation known as convolution to extract information from data objects that are represented in the form of regular grids  (1D vectors, 2D matrices, and high-dimensional tensors) and irregular grids (2D graphs and high-dimensional hypergraphs). These data representations are flexible and can be used to encode a wide range of data objects such as audio signals, chemical spectra, molecules, images, and videos. Moreover, such representations can encode multi-channel data, which allows capturing color and multi-variate inputs (e.g., multi-variate time series and molecular graphs). CNNs use convolution operations to identify features that can be extracted from the data that best explain an output; such features are identified by identifying optimal convolution filters or operators, which are the parameters that are learned by the CNN. The learning process of the operators requires sophisticated optimization algorithms and can be a computationally expensive process.  CNNs is a class of models of an emergent ML field known as geometric deep learning, which leverages tools from geometry and topology to represent and process data. 
\\

The earliest version of a CNN was proposed in 1980 by Kunihiko Fukushima \cite{fukushima1980neocognitron} and was used for pattern recognition. In the late 1980s, the LeNet model proposed by LeCun et al. introduced the concept of \textit{backward propagation}, which streamlined learning computations using optimization techniques \cite{le1989handwritten}. Although the LeNet model had a simple architecture, it was capable of recognizing hand-written digits with high accuracy. In 1998, Rowley et al. proposed a CNN model capable of performing face recognition tasks  (this work revolutionized object classification and detection) \cite{rowley1998neural}. The complexity of CNN models (and their predictive power) has dramatically expanded with the advent of parallel computing architectures such as graphics processing units \cite{nickolls2008scalable}. Modern CNN models for image recognition include SuperVision \cite{krizhevsky2012imagenet}, GoogLeNet \cite{szegedy2014going}, VGG \cite{simonyan2014very}, and ResNet \cite{he2015deep}. New models are currently being developed to perform diverse computer vision tasks such as object detection \cite{ren2015faster}, semantic segmentation \cite{long2015fully}, action recognition \cite{simonyan2014two}, and 3D analysis \cite{ji20123d}. Nowadays, CNNs are routinely used in smartphones (unlock feature based on face recognition) \cite{network2017face}.  
\\

While CNNs were originally developed for computer vision, the grid data representation used by CNNs is flexible and can be used to process datasets arising in many different applications. For instance, in the field of chemistry, Hirohara and co-workers proposed a matrix representations of SMILES strings (which encodes molecular topology) by using a technique known as one-hot encoding \cite{hirohara2018convolutional}. The authors used this representation to train a CNN that could predict  the toxicity of chemicals; it was shown that the CNN outperformed traditional models based on fingerprints (an alternative molecular representation). In biology, Xie and co-workers applied CNNs to count and detect cells from micrographs \cite{xie2018microscopy}.  More recently, CNNs have been expanded to process graph data representations, which has greatly expanded their application scope. These types of CNNs (known as graph neural networks) have been widely used in the context of molecular property predictions \cite{duvenaud2015convolutional,gilmer2017neural}.
\\

Manufacturing covers a broad space of important products and processes that is virtually impossible to enumerate; in this chapter, we focus our attention on applications of CNNs to examples of potential relevance to chemical and biological manufacturing (which cover domains such as pharmaceuticals, agricultural products, food products, consumer products, petrochemicals, and materials).  We also highlight that these manufacturing sectors are seeing an emergent use of autonomous platforms that enable flexible and high-throughput experimentation and/or on-demand production production; as such, the concepts discussed can be applicable in such context. We provide specific case studies that we believe provide representative examples on how CNNs can be used to facilitate decision-making in manufacturing. Specifically, we show how to use CNNs to i) decode multivariate time series data; ii) decode complex signals generated from microscopy and flow cytometry to detect contaminants in air and solution; iii) decode real-time ATR-FTIR spectra to characterize plastic waste streams; iv) predict surfactant properties directly from their molecular structures, and v) map image data into signals for feedback control.  

\section{Data Objects and Mathematical Representations}

A wide range of datasets encountered in manufacturing can be represented in the form of a couple of fundamental mathematical objects: {\em tensors and graphs}. Such representations are so general that, in fact, it is difficult to imagine a dataset that cannot be represented in this way. The key distinction between a tensor and a graph is that a tensor is a regular object (e.g., a regular mesh) while a graph is not (e.g., an irregular mesh). Moreover, tensors implicitly encode positional context, while graphs might or might not encode positional context. Differences between tensors and graph representations play a key role in the way that CNN architectures are designed to extract information from data. Unfortunately, it is often not obvious what data representation is most suitable for a particular application and sometimes the representation might not naturally emerge from the data. In fact, one can think of CNNs as a tool for {\em representation learning}, in the sense that it aims to learn the best way to represent the data to make a prediction.  
 
\subsection{Tensor Representations}

Data objects are often attached to a grid; the most common example of this is a grayscale image, which can be represented as a 2D grid object. Here, every spatial grid point is a pixel and the data entry in such pixel is the intensity of light. A grayscale video is simply a sequence of grayscale images and can be represented as a 3D grid object (a couple of spatial dimensions plus time); here, every space-time grid point is a voxel and the data entry is the intensity of light.  A common misconception of grid data is that it can only be used to represent images and videos but its scope is much broader. For instance,  3D density fields of chemicals or velocities (as those generated using molecular and fluid dynamics simulations) can be represented as a 3D grid; moreover, the thermal field of a surface can be represented as a 2D grid. 
\\

Tensors are mathematical objects used to represent grid data. A tensor is a generalization of vectors (1D tensors) and matrices (2D tensors) to high dimensions \cite{goodfellow2016deep}. A key property of a tensor is that it implicitly encodes positional context and order (it lives in a Euclidean space); specifically, every entry of a tensor has an associated set of coordinates/indexes that uniquely specify the location of an entry in the tensor. Due to its positional context, the nature of the tensor can  altered by rotations; for instance, rotating an image (or transposing its associated matrix) distorts its properties.  
\\

\begin{figure}[!htb]
  \centering
      \includegraphics[width=\textwidth]{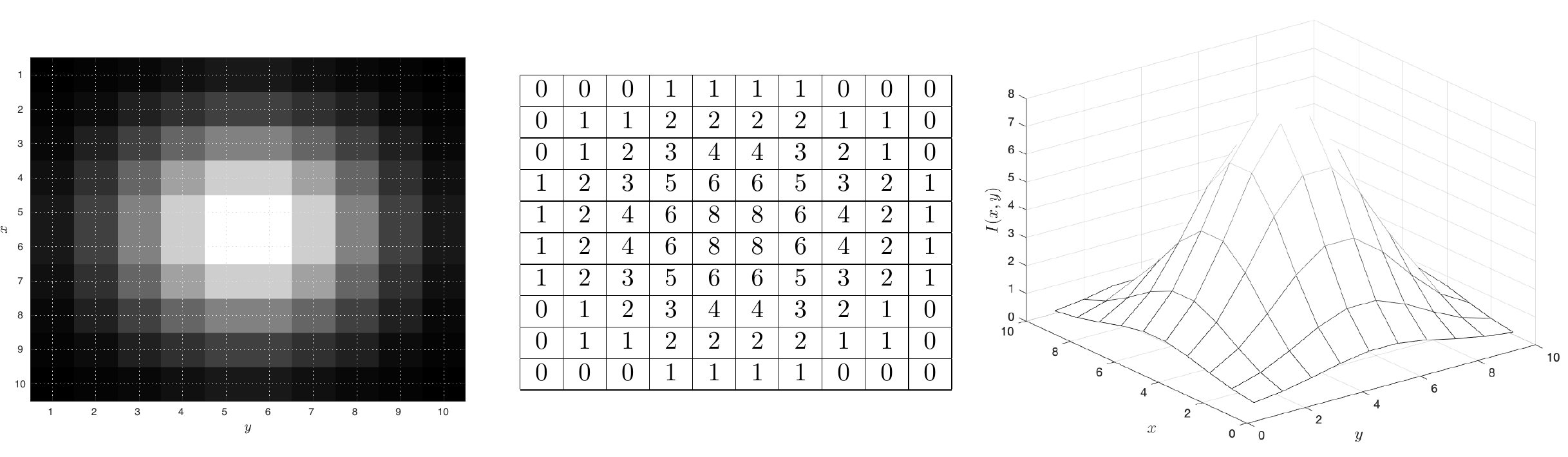}
  \caption{Representation of a grayscale image (left) as a 2D grid object (middle). The grid is a matrix in which each entry represents a pixel and the numerical value in the entry is the intensity of light. Representation of the grayscale image as a manifold (right); this reveals geometrical patterns of the image.} \label{fig:grid}
\end{figure}

Tensors are flexible objects that can also be used to represent multi-attribute/multi-channel grid data. For example, color images can be represented as a superposition of three grids  (red, green, and blue channels). Here, each channel is a 2D tensor (a matrix) and the stacking of these three channels is a 3D tensor.  Channels can also be used to represent multi-variate data in each entry of a grid. For instance, an audio or sensor signal is a time series that can be represented as a one-channel vector, while a multivariate time series (e.g., as such obtained from a collection of sensors in a manufacturing facility) can be represented as multi-channel vector. 
\\

It is important to highlight that there is an inherent duality between images (reality as perceived by human vision or an optical device) and tensors (its mathematical representation). Specifically, images are optical fields that our visual sensing system capture and process to navigate the world and make decisions, while tensors are artificial mathematical representations used for computer processing. Making this distinction explicit is important, because humans typically excel at extracting information from visual data (without having any knowledge of mathematics) compared to numerical data (e.g., number sequences).  As such, this begs the questions: {\em Why do automation systems present data to human operators as numbers?  What are the best visual representations that humans can use to interpret and analyze data more easily?}  These questions are at the core of human-computer interaction and highlight the relevance of data visualization and processing techniques. 
\\

It is also important to highlight that the human vision system and the brain have inherent limitations in sensing and interpreting optical fields. For instance, the human vision and auditory system cannot capture all frequencies present in an optical field or an audio signal; as such, we need instrumentation (e.g., microscopes and nocturnal vision systems) that reveal/highlight information that cannot be captured with our limited senes. Moreover, the human brain often gets ``confused" by distortions of optical fields and audio signals (e.g., rotations and deformations) and by noise (e.g., fog and white noise). These limitations can be overcome with the use of artificial intelligence tools such as CNNs.  Here, sensing signals such as images and audio signals are represented mathematically as grid data to extract information. Unfortunately, grid data representations are limited in that they inherently represent regular objects and are susceptible to rotation and deformations. 

\subsection{Graph Representations}

Graphs provide another flexible and powerful mathematical data representation. A graph is a topological object that consists of a set of nodes and a set of edges; each node is a point in a graph object and each edge connects a pair of nodes. The connectivity (topology) of a graph is represented as an adjacency matrix. 
\\

\begin{figure}[!htb]
  \centering
      \includegraphics[width=\textwidth]{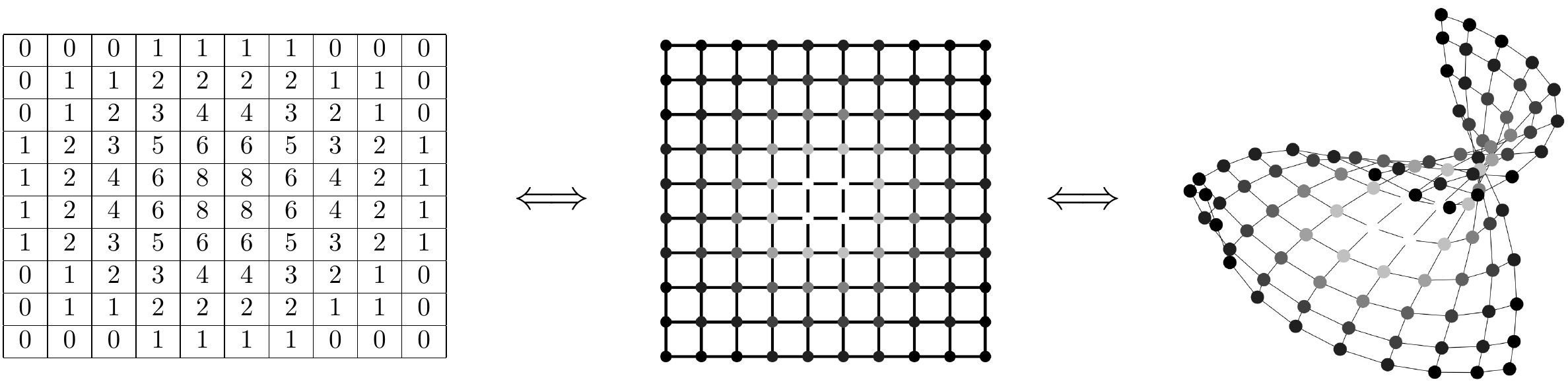}
  \caption{Representation of a grayscale image (left) as graph (middle). Each node in the graph represents a pixel and the weight in the node encodes the intensity of light. The graph is a topologically invariant object that is not affected by deformations (right).} \label{fig:graph}
\end{figure}

\begin{figure}[!htb]
  \centering
      \includegraphics[width=0.3\textwidth]{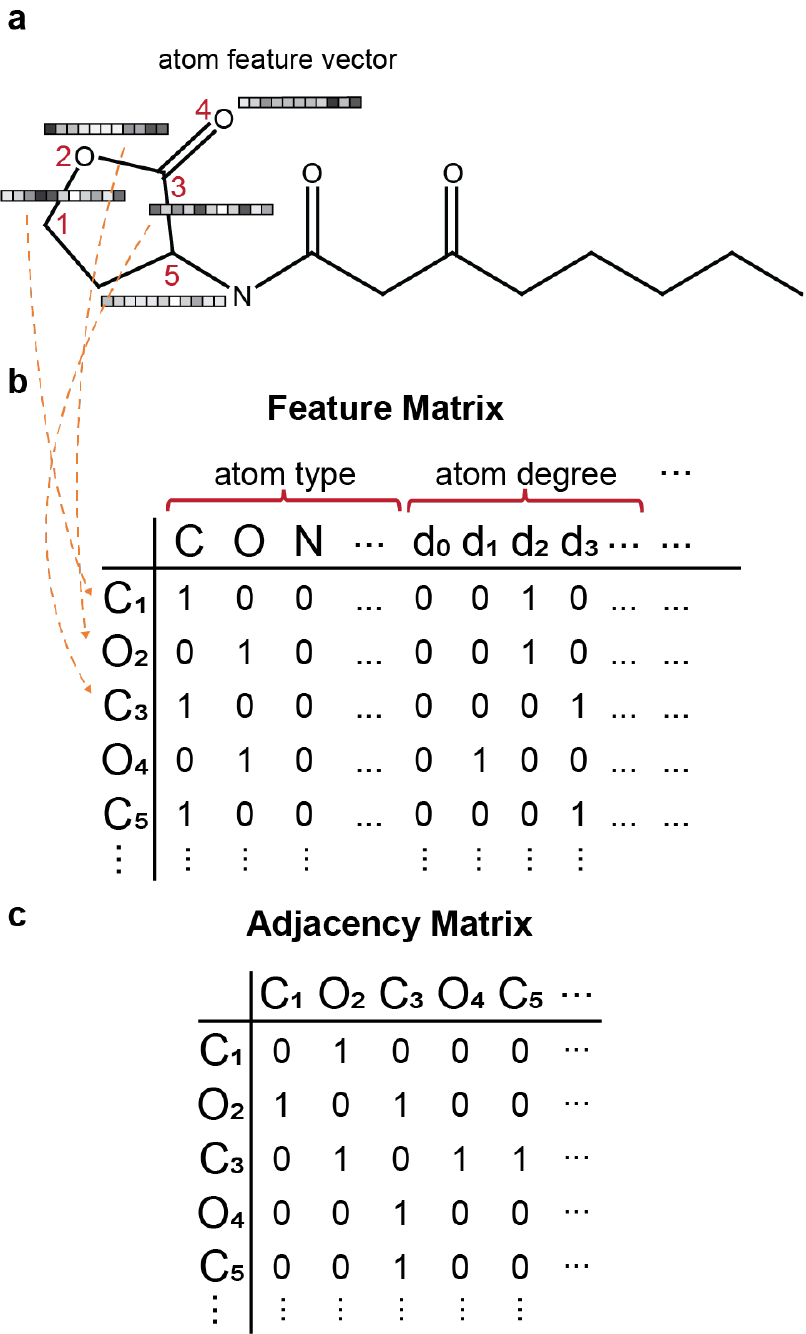}
  \caption{Representation of a molecule as a multi-channel/multi-attribute graph. The molecule topology is encoded in the adjacency matrix, while features of the atoms are encoded as vectors that are stacked in a feature matrix. Reproduced from \cite{qin2021predicting} with permission from the American Chemical Society.} \label{fig:molecular_graph}
\end{figure}

A graph defines an irregular grid and one can attach multi-channel data (features) to nodes and edges in such grid. For instance, when representing a molecule, one can attach multiple features to each atom (e.g., identity and charge). Here, one can think of each feature as a the channel of the graph; graphs with multiple features in nodes and edges are also known as multi-attribute graphs. An example on how multi-attribute graphs are used to represent molecules is presented in Figure \ref{fig:molecular_graph}. 
\\

A fundamental property of a graph is that it does not directly encode positional context (it does not live in a Euclidean space). As such, the graph can be stretched and rotated or the nodes can be shuffled around in space without affecting the underlying topology of the graph. In other words, the topology of the graph is fully dictated by the node-edge  connectivity. This gives graph representations many useful properties (such as rotational invariance), which can be highly beneficial when representing certain systems. For instance, in molecular simulations, molecules might move around randomly in space, but their connectivity (e.g., hydrogen bonds) might not be affected \cite{je2022integration}.  Another key property of graphs is that, because these are irregular grid objects, one can develop CNNs that learn features of graphs of different sizes (e.g., polymer and peptide sequences). 
\\

It is worth highlighting that, while graphs are inherently 2D objects with no positional context,  it is possible to encode 3D (or higher dimensional) position to each node in a graph as features \cite{yang2019analyzing}. For instance, in a graph that represents the time evolution of a supply chain network, node features can contain the specific time point and the geographical coordinates. Moreover, it is possible to generalize graph representations to hypergraphs, which allow representing objects in which an edge connects multiple nodes.  These more advanced geometric representations give rise to an emerging field of machine learning known as geometric deep learning \cite{bronstein2017geometric} and to topological data analysis \cite{smith2021topological,smith2021euler}. These approaches enable the analysis of complex 3D (and higher-dimensional) objects.

\subsection{Color Representations}

\begin{figure}[!htb]
  \centering
      \includegraphics[width=0.7\textwidth]{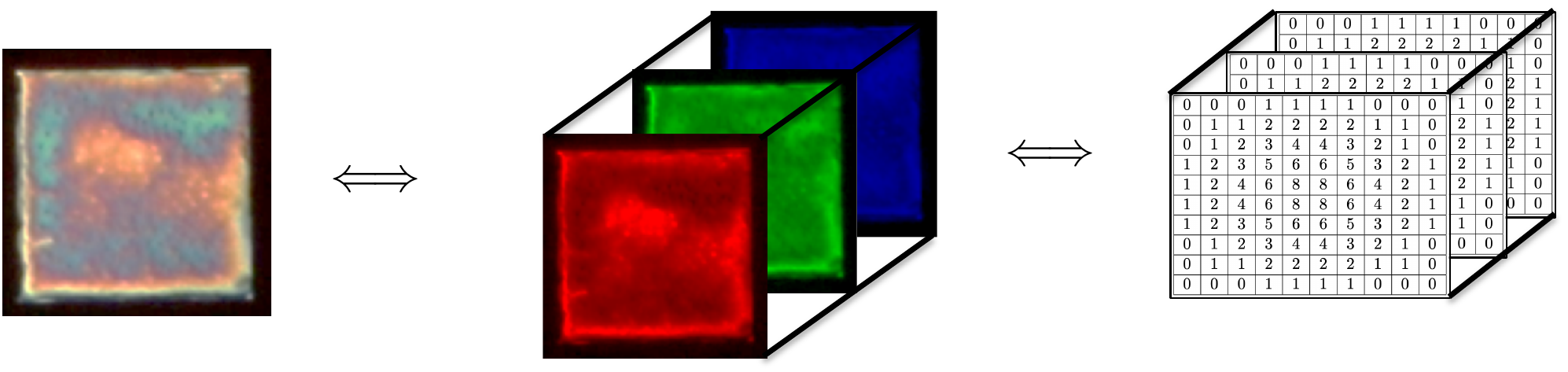}
  \caption{Representation of a color image (left) as a superposition of red, green, and blue channels (middle). Each channel is a 2D grid object (a matrix) and stacking the channels generates a 3D tensor.} \label{fig:color}
\end{figure}

The human vision system uses color as a key source of information to make decisions. For computer processing, color is typically represented using a red-green-blue (RGB) color space (see Figure \ref{fig:color}). The RGB color space is an additive space in which the red, green, and blue channels are mixed together to produce a wide range of colors \cite{hirsch2004exploring}. 
\\

Before the RGB space became the default in computer electronic systems, it already had a solid theory based on human perception of color. The choice of these three primary colors is related to the physiology of the human eye; specifically, the three photoreceptor cells (cone cells) in the human eye have peak sensitivity at three wavelengths associated red, green, and blue. The RGB color space was designed to maximize the difference in response of cone cells to different wavelengths of light \cite{hunt2005reproduction}.  However, there are many other color spaces that can be useful in different situations. For example, the CIEL*A*B (or LAB) color space is a powerful representation that can be used for quantitative color comparisons. 
\\

The LAB space was defined by the International Commission on Illumination (CIE) to be a perceptually uniform color space, meaning that a group of colors separated by the same distance in LAB color space will look equally different. The three coordinates of the LAB color space represent the brightness of the color ($L*=0$ for black, $L*=100$ for white), its position between red and green ($a$, negative values indicate green, positive values indicate red), and its position between yellow and blue ($b$, negative values indicate blue, positive values indicate yellow). One can convert the RGB color space to a LAB color space using a nontrivial, nonlinear transformation. The LAB space can be used to highlight image features that might not be directly obvious to the human eye; this is a principle that can be exploited by CNNs to extract hidden information from images. 
\\

Unlike RGB and LAB imaging, which captures only three wavelength bands in the visible spectrum, spectral imaging can use multiple bands across the electromagnetic spectrum \cite{chang2003hyperspectral}. Spectral imaging can use infrared, visible spectrum, ultraviolet, X-rays, or some combination of the above. Multispectral and hyperspectral cameras can capture hundreds of bands for each pixel in an image, which can be interpreted as a complete spectrum. Due to the wealth of information encoded in spectral images, this technique has been widely used in agriculture \cite{lu2020recent}, healthcare \cite{fei2020hyperspectral}. and non-destructive analysis of materials \cite{manley2014near}. 
 
\section{CNN Architectures}

CNNs denote a broad class of ML models that utilize specialized convolutional blocks to extract feature information from data objects. CNNs were originally designed to extract data from grid objects but they recently have been adapted to extract information from graphs.  The key mathematical operation behind CNNs is convolution; loosely speaking, this operation transforms the data in a domain of the data object by applying a weighted sum of the entries in such domain. The weights of such sum are defined by a convolution filter or operator. The goal of the CNN is to identify optimal filters or operators that maximize information extracted from a collection of input data objects; here, ``maximum information" means that such information can predict outcomes (labels) associated with the input data objects.  For instance, one can train a CNN to predict the toxicity of a molecule directly from its molecular structure; this is done by training a CNN with a collection of molecules and associated toxicity levels.  
\\

\begin{figure}[!htb]
  \centering
      \includegraphics[width=\textwidth]{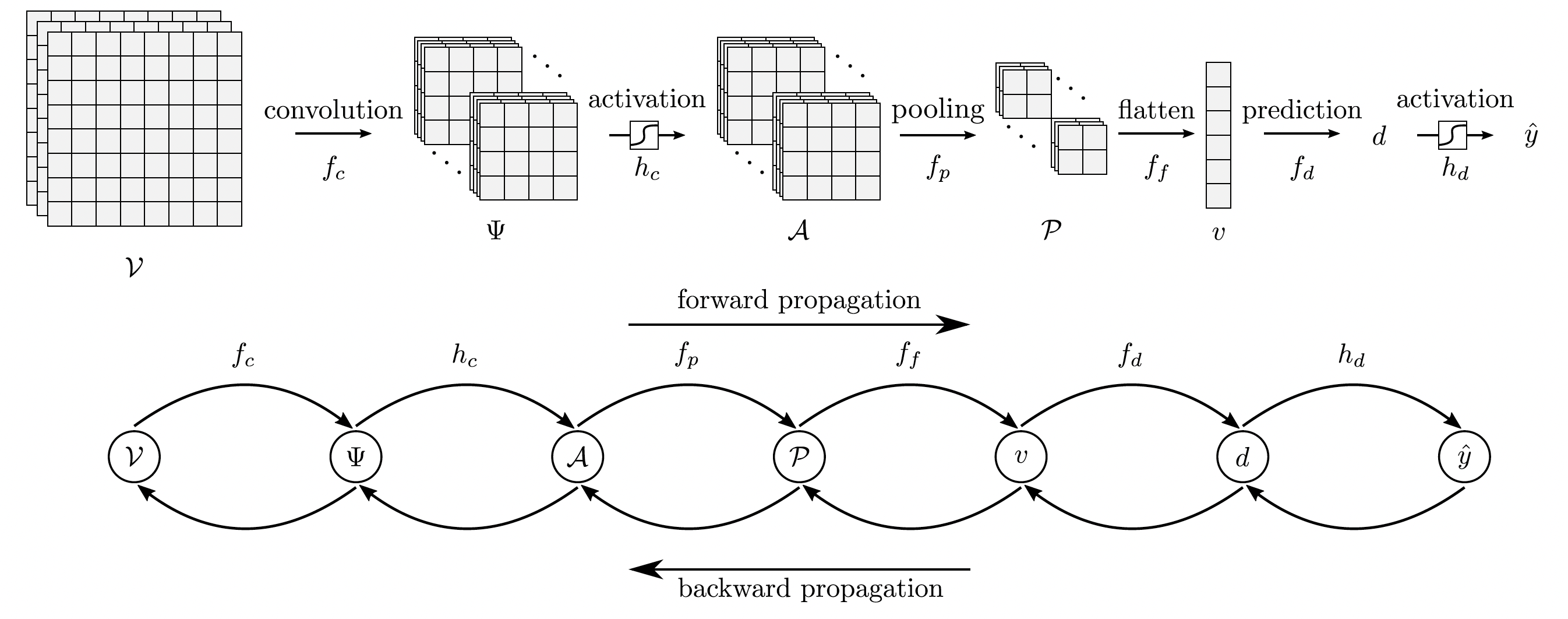}
  \caption{High-level view of a CNN architecture. The input data object (e.g., grid or graph) is propagated through a sequence of convolution, activation, and pooling operations to make an output prediction. The trainable parameters of the operations are learned by minimizing a loss function. The derivative/gradient of the loss function is computed using backpropagation. Reproduced from \cite{jiang2021convolutional} with permission from Wiley publishing.} \label{fig:cnn}
\end{figure}

A CNN architecture is simply a forward propagation of the input data object through a sequence of convolution, activation, and pooling operations that contain parameters that are learned by minimizing a loss function. A backward propagation (backpropagation) scheme is used to compute the gradient of the loss function with respect to the parameters. 
\\

In this section, we present a basic discussion on the mathematics of convolution and computational procedures behind the training of CNNs.  With this, we aim to highlight the inner workings of CNNs and associated bottlenecks and challenges. 

\subsection{Convolution Operations} 

\begin{figure}[!htb]
  \centering
      \includegraphics[width=0.7\textwidth]{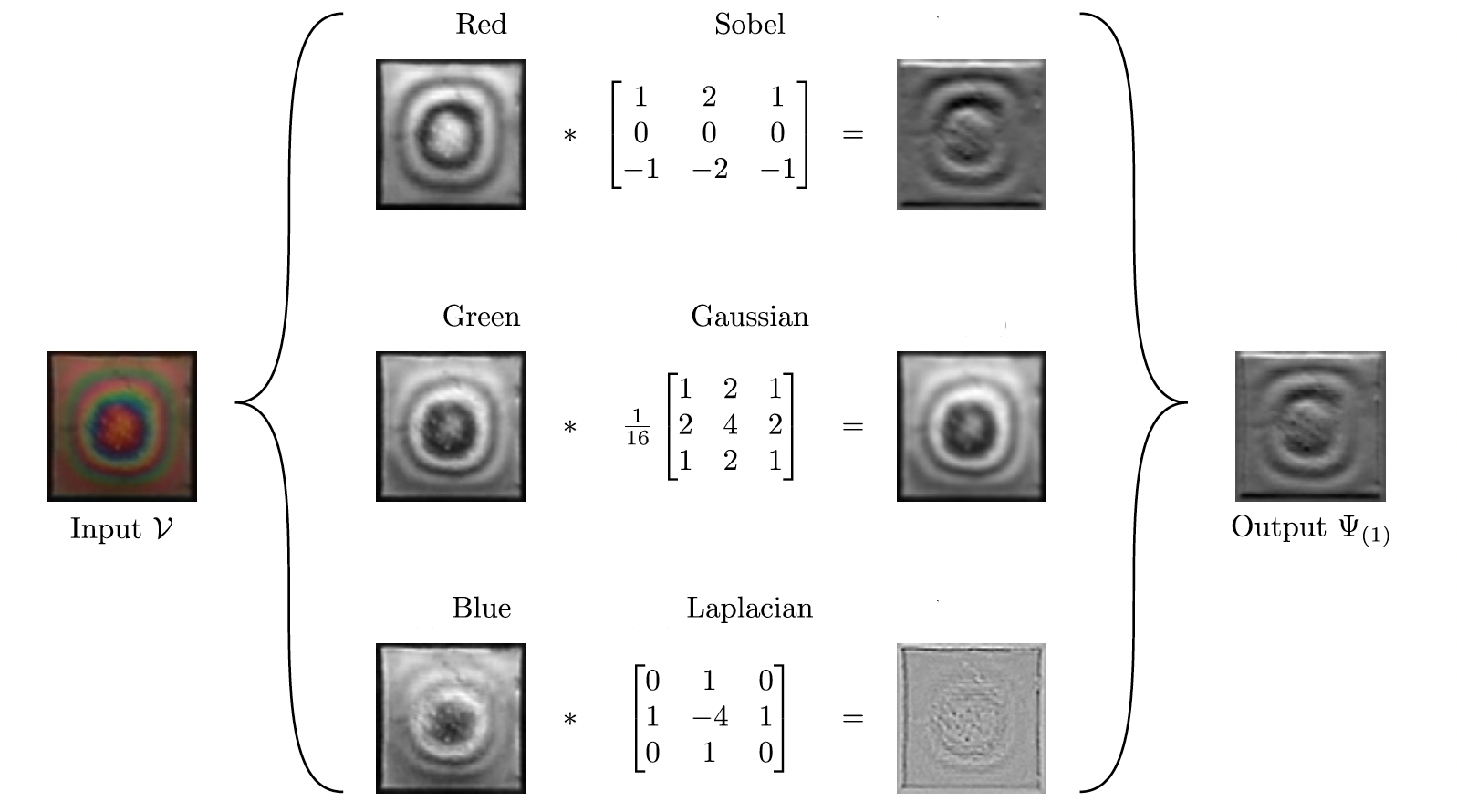}
  \caption{Convolution of a 3-channel grid (a color image). Each channel is processed using a different convolution operator. Each operator highlights/extracts different features (e.g., geometrical features) from the image. Reproduced from \cite{jiang2021convolutional} with permission from Wiley publishing.} \label{fig:convolutions}
\end{figure}

We will first consider CNN networks $f_{\text{cnn}}$ that map an input tensor $V$ to output predictions $\hat{y}$. A particular convolutional layer takes an input object $V$ and \emph{convolves} it by applying a convolutional operator $U$ which employs a set of convolutional filters or operators to produce a feature map $\Psi$. The convolution operation is denoted as $\Psi = U * V$ and is simply a weighted summation between the entries of the convolution operator and the input object. Here, we also notice that the operator is typically much smaller than the input object and thus the operator is applied over a moving window (neighborhood) that covers the entire input object. This operation is not well-defined by the boundaries of $V$, as some indexing will violate the input domain, but this is typically resolved by adding padding of zero-valued entries around the image (a technique called zero-padding). We can also express this operation in the compact form $\Psi = f_c(V; U)$.  As shown in Figure \ref{fig:convolutions}, different operators highlight different features of the input object.  Moreover, we note that one can propose a wide range of operators to extract information from the input data. In a CNN architecture, these operators are learned in a way that they extract information that best explains an output of interest. 

Following Figure \ref{fig:pattern}, a common and intuitive interpretation of applying convolution operators is in terms of pattern recognition. The filters that comprise $U$ each embed a particular pattern that, when convolved with a particular neighborhood of $V$ (as depicted in Figure \ref{fig:convolutions}), gives a score of how well the pattern is matched (where larger values denote greater matching). Hence, the feature map $\Psi$ can be interpreted as a scoresheet of how certain patterns (as encoded by the filters in $U$) are manifested in $V$. 
\\

\begin{figure}[!htb]
  \centering
      \includegraphics[width=0.6\textwidth]{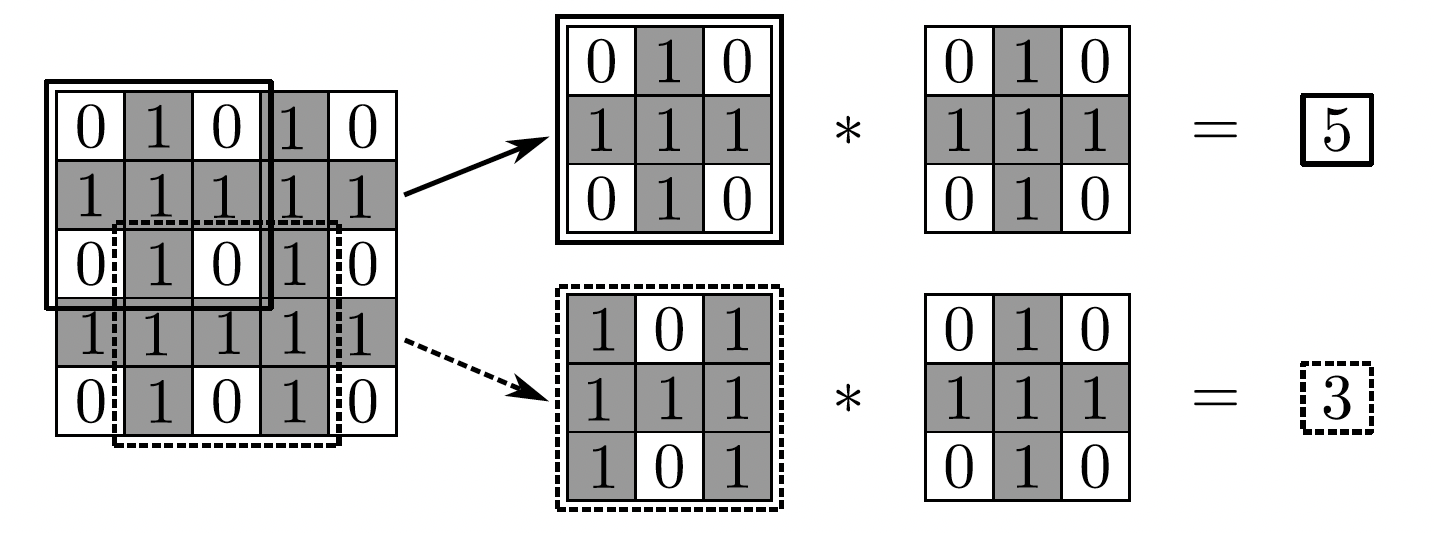}
  \caption{Convolution as a pattern matching technique. The convolution operator highlights regions in the grid in which the patterns of the image and of the operator matched. Reproduced from \cite{jiang2021convolutional} with permission from Wiley publishing.} \label{fig:pattern}
\end{figure}

In a graph CNN architecture \cite{zhou2020graph}, the input graph is propagated based on its topology; here, convolution operations are performed analogously to grid data objects, by using weighted summations of the features of a node and of its neighboring nodes. A  message passing framework \cite{gilmer2017neural} is typically used as a generalized approach to build GNN architectures that incorporate both node and edge features.  
  
\subsection{Activation Functions}

The feature map output of a convolutional layer is typically mapped element-wise through an activation function $\alpha$ to yield the activated object $A$. This operation can be written as $A = h_c(\Psi)$ and helps the CNN encode nonlinear behavior. Common choices for the activation function include:
\begin{equation*}
    \begin{gathered}
        \alpha_\text{sig}(z) = \frac{1}{1 + e^{-z}} \\
        \alpha_\text{tanh}(z) = \text{tanh}(z) \\ 
        \alpha_\text{ReLU}(z) = \text{max}(0, z).
    \end{gathered}
\end{equation*}
Here, the Rectified Linear Unit (ReLU) function $\alpha_\text{ReLU}(\cdot)$ is highly popular since it generally exhibits greater sensitivity to changes in the input \cite{nair2010rectified}. Moreover, this function can be easily represented using piecewise linear functions and mixed-integer programming formulations. 

\subsection{Pooling}

Another key component of CNNs are pooling layers; pooling operations are dimension reduction mappings $f_p$ that seek to summarize/reduce the activation $A$ by collapsing certain sub-regions of smaller dimension (referred to as pooling). Pooling helps to make the learned representation more invariant to small length-scale perturbations \cite{nagi2011max}. Common choices include max-pooling and average-pooling where the maximum or average of a sub-region is used to scalarize, respectively. 

\subsection{Convolution Blocks}

Convolution blocks in CNNs denote the combination of convolution, activation, and pooling of a given input. In other words, a convolution block employs the mapping $f_{\text{cb}}$ such that:
\begin{equation}
    P = f_{\text{cb}}(V; U) = f_p(h_c(f_c(V; U))).
    \label{eq:convolution_block}
\end{equation}
Blocks can employ more complex operation nesting (e.g., multiple convolutional layers), but we consider blocks that follow \eqref{eq:convolution_block} for simplicity in presentation. Moreover, these can take pooled outputs $P$ as input and thus facilitate the use of multiple convolutional blocks in succession via recursively calling $f_{\text{cb}}$. These blocks essentially act as a feature extractors via convolutional filters they employ. Moreover, this feature extraction becomes more specialized for blocks that are located deeper in the CNN. In the context of computer vision, this means that the first blocks extract simple features (e.g., edges or colors) and the deeper blocks can extract more complex patterns (e.g., particular shapes). Thus, we can think of performing multiple convolution operations as a {\em multi-stage distillation process} where the successive feature spaces capture patterns of increased length-scale and complexity (i.e., distill the image data into an increasingly purified features) \cite{chollet2017deep}. 

\subsection{Feedforward Neural Networks}

The output $P$ of the last convolution block is typically flattened (vectorized); this can be represented as the mapping $f_f$ and yields the feature vector $v$:
\begin{equation}
    v = f_f(P).
\end{equation}
The feature vector is then fed into a feedforward neural network model $f_d$ which predicts the desired state space vector $\hat{y}$. Figure \ref{fig:cnn} illustrates a typical image CNN model that implements the components described above. It employs a couple of convolution blocks and can be described in the functional form:
\begin{equation}
    \hat{y} = f_{\text{cnn}}(V),
\end{equation}
where the mapping $f_{\text{cnn}}$ is a nested set of convolution, activation, pooling, and flattening operations. This emphasizes that convolution blocks are feature information extractors and the dense layers act as the predictors whose feature space is the flattened output $v$ of the final block.

\subsection{Data Augmentation}

\begin{figure}[!htb]
  \centering
  \begin{subfigure}[b]{0.2\textwidth}
      \centering
      \includegraphics[width=\textwidth]{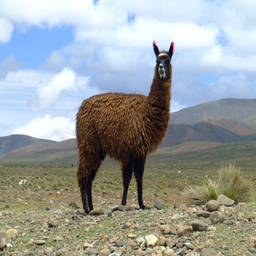}
      \caption{Original}
  \end{subfigure}
  \quad
  \begin{subfigure}[b]{0.2\textwidth}
      \centering
      \includegraphics[width=\textwidth]{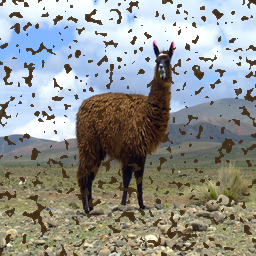}
      \caption{Splattered}
  \end{subfigure}
  \quad
  \begin{subfigure}[b]{0.2\textwidth}
      \centering
      \includegraphics[width=\textwidth]{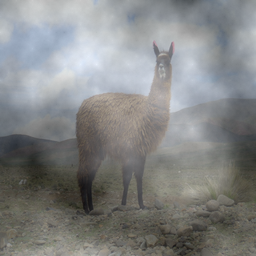}
      \caption{Fogged}
  \end{subfigure}
  \quad
  \begin{subfigure}[b]{0.2\textwidth}
      \centering
      \includegraphics[width=\textwidth]{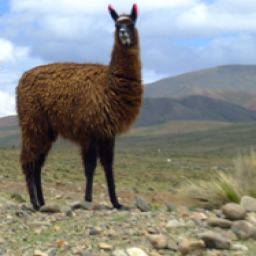}
      \caption{Shifted}
  \end{subfigure}
 \caption{Examples of image perturbation methods used in data augmentation techniques. Reproduced from \cite{pulsipher2022safe} with permission from Elsevier.}
 \label{fig:image_aug}
\end{figure}

Data augmentation denotes a class of techniques that aim to artificially expand the size of a training dataset by applying varied perturbations/transformations to the input data.  In the context of computer vision, augmentation is often used to expand the size of the training image set in an effort to decrease the likelihood of the CNN encountering novel images and in an effort to avoid rotation invariance issues associated with grid data objects. Image augmentation generally denotes perturbing the training images such that the CNN sensor can be robust to those types of visual disturbance. Common perturbations include rotation, translation, cropping, blurring, brightness changing, splattering, and more. There are many software tools available that implement these transformations, which include \texttt{TensorFlow} and \texttt{ImgAug} \cite{geron2019hands, imgaug}. Figure \ref{fig:image_aug} shows how a training image is augmented via a variety of perturbations. This methodology helps mitigate the risk of CNNs encountering novel images, but it is not typically possible to account for all the disturbance a process might encounter.

\subsection{Training and Testing Procedures}

Training CNN procedures seek optimal model parameters (i.e., convolution operators and dense network weights) that minimize the error incurred by the output CNN predictions relative to the outputs of the training dataset. Here, we consider a training set $|\mathcal{K}|$ that contains input-output pairs. The prediction error minimized in the training procedure is called the loss function $L$. For example, regression models typically use a sum-of-squared-error (SSE) loss function:
\begin{equation}
    L(\hat{y}) = ||\hat{y} - y||_2^2.
\end{equation}
Thus, by grouping all the CNN model parameters into $\theta$ we can express model training as a standard optimization problem:
\begin{equation}
    \begin{aligned}
        &\min_\theta &&  \sum_{k \in \mathcal{K}} L(\hat{y}^{(k)}) \\
        &\text{s.t.} && \hat{y}^{(k)} = f_{\text{cnn}}(V^{(k)}; \theta), && k \in \mathcal{K}.
    \end{aligned}
\end{equation}
This can readily be expressed as an unconstrained optimization problem by inserting the constraint equations directly into the objective. Stochastic Gradient Descent (SGD) is typically used to solve this problem due to the large amount of training data, the high number of model parameters, and the model complexity. Moreover, forward and backward propagation techniques are used to evaluate the objective and derivative values required by each iteration of the SGD algorithm (see Figure \ref{fig:cnn}). For classification models, one typically uses an entropy loss function. 

\subsection{CNN Architecture Optimization}

Neural architecture search (NAS) has been designed to automatically search for the best CNN architecture for a given dataset. Specifically, it uses a search method  (e.g., reinforcement learning, evolutionary algorithms, or stochastic gradient descent) to explore the user-defined search space and chooses the best architecture based on the performance of the generated model (e.g., validation accuracy) on a given task. The search space contains all possible architectures. The architectures discovered by NAS have been shown to outperform manually designed architectures in various tasks such as image classification \cite{real2017large}, image segmentation \cite{liu2019auto}, natural language processing \cite{fan2020searching}, and time series prediction \cite{maulik2020recurrent}.

\subsection{Transfer Learning} 

The training process of a CNN can be highly computationally intensive because this requires performing a large number of convolution operations (e.g., at each SGD iteration and for a large training set). However, it is possible to use the filters/operators of existing, trained CNNs to extract information for a new dataset and such information can be used for different tasks (e.g., develop a different ML model such as a support vector machine).  This approach is known as transfer learning and is based on the observation that, while the filters used are not optimal for the new dataset at hand, they can still extract valuable information. For instance, in the context of human learning, one can often detect features for objects that we have not seen before. 

\section{Case Studies}

In this section we present a set of case studies that highlight potential uses of CNNs in a manufacturing context. Our goal with this is not to provide an exhaustive list of applications, but rather to highlight the capabilities of CNNs as well as to show how data representations can be used in creative ways to tackle diverse problems. 

\subsection{CNNs for Sensor Design}

\subsubsection{Detecting Contaminants in Solution}

\begin{figure}[!htb]
	\begin{center}
		\includegraphics[width=0.7\linewidth]{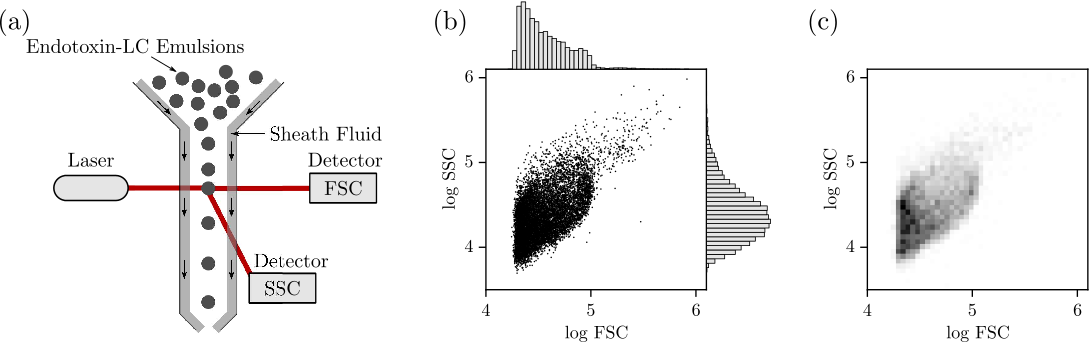}
		\caption{Overview of the interaction between an endotoxin and LC emulsions. (a) Generation of FSC/SSC scatter fields. The endotoxin-LC emulsions are pumped into the flow cytometer in the direction of the sheath flow. Laser light is scattered from the LC droplets and collected at two angles (FSC and SSC). By combining the FSC and SSC data points for 10,000 LC droplets, we generate an FSC/SSC field. (b) Scatter field generated by LC droplets exposed to 100 pg/mL of endotoxin. Marginal probability densities of FSC (top) and SSC (right) light in log scale are generated with 50 bins. (c) 2D grid of the scatter fields by binning and counting the number of events in a bin. Reproduced from \cite{jiang2021using} with permission from the Royal Society of Chemistry.}
		\label{fig:endo1}
	\end{center}
\end{figure}

This case study is based on the results presented in \cite{jiang2021using}.  Endotoxins are lipopolysaccharides (LPS) present in the outer membrane of bacteria \cite{Borek1985}. Recently, it has been observed that micrometer-sized liquid crystal (LC) droplets dispersed in an aqueous solution can be used as a sensing method to detect and measure the concentration of endotoxins of different bacterial organisms.  After exposure to endotoxins, LC droplets undergo transitions that yield distinct optical signatures that can be quantified using flow cytometry (Figure~\ref{fig:endo1}). 

Flow cytometry produces complex data objects in the form of scatter point clouds of forward and side scattering (FSC/SSC); here, each point represents a scattering event of a given droplet. The key observation is that a point could be converted into a 2D grid data object via binning; this is done by discretizing the FSC/SSC domain and by counting the number of points in a bin. The 2D grid object obtained is a matrix that we can visualize as a grayscale image; here, each pixel is a bin and the intensity is the number of events/droplets in the bin (bins with more droplets appear darker). This visualization is a 2D projection of a 3D histogram (the third dimension corresponds to the number of events, also known as the frequency). In other words, the 2D grid object captures the geometry (shape) of the joint probability density of the FSC/SSC. These geometrical features can be extracted automatically using a CNN.  

\begin{figure}[!htb]
	\begin{center}
		\includegraphics[width=0.4\linewidth]{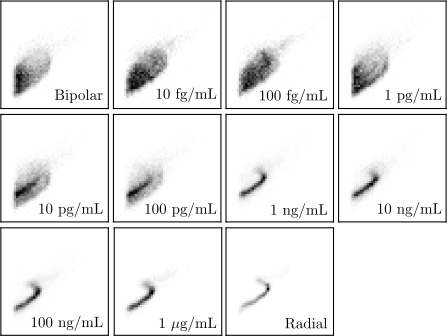}
		\caption{Effect of endotoxin concentration on the FSC/SSC scatter field (represented as 2D grid objects). Scatter fields obtained using LC droplets exposed to different concentrations of endotoxin. As the endotoxin concentration increases, the LC droplet population shifts from a bipolar to a radial configuration. Reproduced from \cite{jiang2021using} with permission from the Royal Society of Chemistry.}
		\label{fig:endo2}
	\end{center}
\end{figure}

\begin{figure}[!htb]
	\begin{center}
		\includegraphics[width=0.7\linewidth]{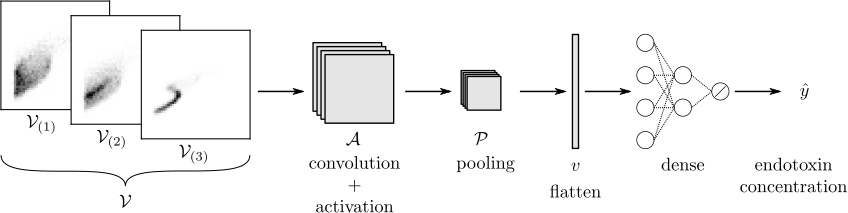}
		\caption{EndoNet architecture. The input to EndoNet is a 3-channel object $\mathcal{V} \in \mathbb{R}^{50 \times 50 \times 3}$; the channels correspond to the target, the negative reference, and the positive reference (each is a matrix of dimension $50 \times 50$). EndoNet includes a convolutional block with 64, $3 \times 3$ operators and a max-pooling block with $2 \times 2$ operators. The feature map $\mathcal{A}$ generated by the convolution and activation blocks is a tensor $\mathbb{R}^{48 \times 48 \times 64}$. The max-pooling block generates a feature map $\mathcal{P} \in \mathbb{R}^{24 \times 24 \times 64}$ that is flattened into a long vector $v \in \mathbb{R}^{36863}$. This vector is passed to two dense layers (each having 32 hidden units). The predicted endotoxin concentration $\hat{y}$ is the output from the dense layer activated by a linear function. Reproduced from \cite{jiang2021using} with permission from the Royal Society of Chemistry.}
		\label{fig:endo3}
	\end{center}
\end{figure}

The effect of endotoxin concentration on the FSC/SSC scatter fields (after binning) is shown in Figure~\ref{fig:endo2}; clear differences in the patterns can be observed at  concentrations that are far apart but differences are subtle at close concentrations.  We trained a 2D CNN that can automatically detect these changes and predict concentrations from the scatter fields.  We used the following procedure to obtain the 2D grid data objects (samples) that were fed to the CNN. For each sample, we generated bins for a given scatter field by partitioning the FSC and SSC dimensions into 50 segments (the grid has $50 \times 50=2,500$ pixels). For each sample, we were also given reference scatter fields that represented limiting behavior: bipolar control (negative) and radial control (positive). Each sample is given by a 3-channel object $\mathcal{V}$  where the first channel is the negative reference matrix $\mathcal{V}_{(1)}$, the second channel is the target matrix $\mathcal{V}_{(2)}$, and the third channel is a positive reference matrix $\mathcal{V}_{(3)}$ (each channel contains a $50\times 50$ matrix). This procedure is illustrated in Figure \ref{fig:endo3}. This multi-channel data representation approach magnifies differences in the target matrix from the references (we will see that neglecting the negative/positive references does not give accurate predictions).  

\begin{figure}[!htb]
	\begin{center}
		\includegraphics[width=0.3\linewidth]{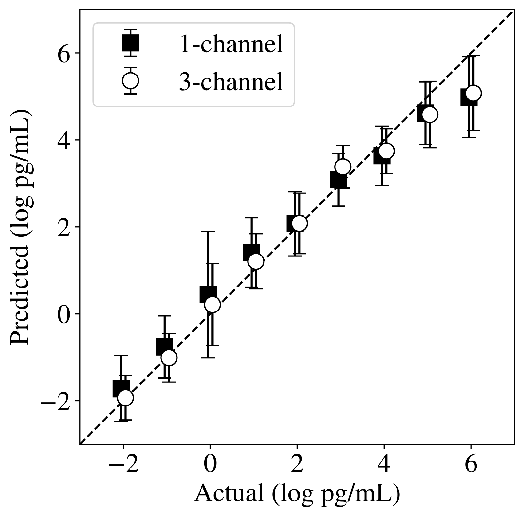}
		\caption{Predicted and actual concentrations at different concentrations using 1-channel and 3-channel representations. Reproduced from \cite{jiang2021using} with permission from the Royal Society of Chemistry.}
		\label{fig:endo4}
	\end{center}
\end{figure}

The 3-channel data object was fed to a CNN, which we call {\em EndoNet}. EndoNet has an architecture of Conv(64)-MaxPool-Flatten-Dense(32)-Dense(32)-Dense(1). The output block generates a scalar prediction $\hat{y}$ (corresponding to the endotoxin concentration). In other words, the CNN seeks to predict the endotoxin concentration from the input flow cytometry fields. The architecture of EndoNet is shown in Figure~\ref{fig:endo3}. The regression results for EndoNet are presented in Figure \ref{fig:endo4}. EndoNet extracts pattern information within and between each channel of the input image. Capturing differences between channels provides context for the CNN and has the effect of highlighting the domains in the scatter field that contain the most information. To validate this hypothesis, we conducted predictions for EndoNet using the 1-channel representation as input (we ignored the positive and negative channels). For the 1-channel representation, we obtained an RMSE of 0.97; for the 3-channel representation we obtained and RMSE of 0.78. It is particularly remarkable that EndoNet can accurately predict concentrations that span eight orders of magnitude. 

\subsubsection{Detecting Contaminants in Air}

\begin{figure}[!htp] 
   \centering
   \includegraphics[width=0.6\textwidth]{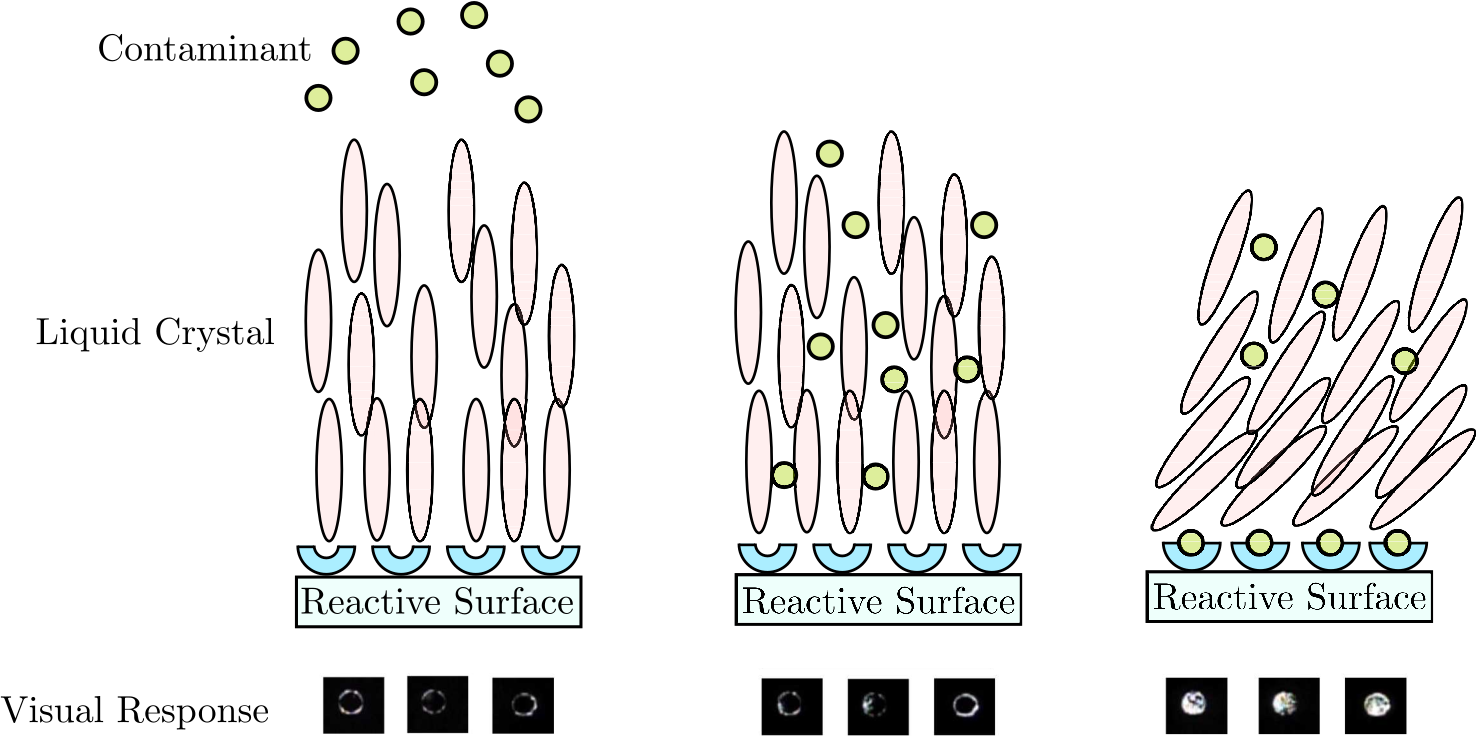} 
   \caption{Working design principles of a liquid crystal chemical sensor. A functionalized LC film selectively responds to the presence of an air contaminant and triggers an optical response. Reproduced from \cite{smith2020convolutional} with permission from the American Chemical Society.}
   \label{fig:LC1}
\end{figure}

This case study is based on work presented in \cite{smith2020convolutional}. LCs also provide a versatile platform for sensing of air contaminants \cite{shah2001principles, mulder2014chiral} and for sensing of heat transfer and shear stress (mechanical sensing) \cite{ireland2000liquid}. In the context of air chemical sensing, an LC sensor can be prepared by supporting a thin LC film on a chemically functionalized surface. The molecules within the LC film (the mesogen) bind to the surface and assume a homeotropic orientation that provides an initial optical signal. Subsequent exposure of the LC film to an analyte leads to diffusive transport of the analyte through the LC phase and displacement of the mesogen at the surface, triggering rich space-time optical responses (see Figure \ref{fig:LC1}). 

\begin{figure}[ht] 
   \centering
   \includegraphics[width=0.5\textwidth]{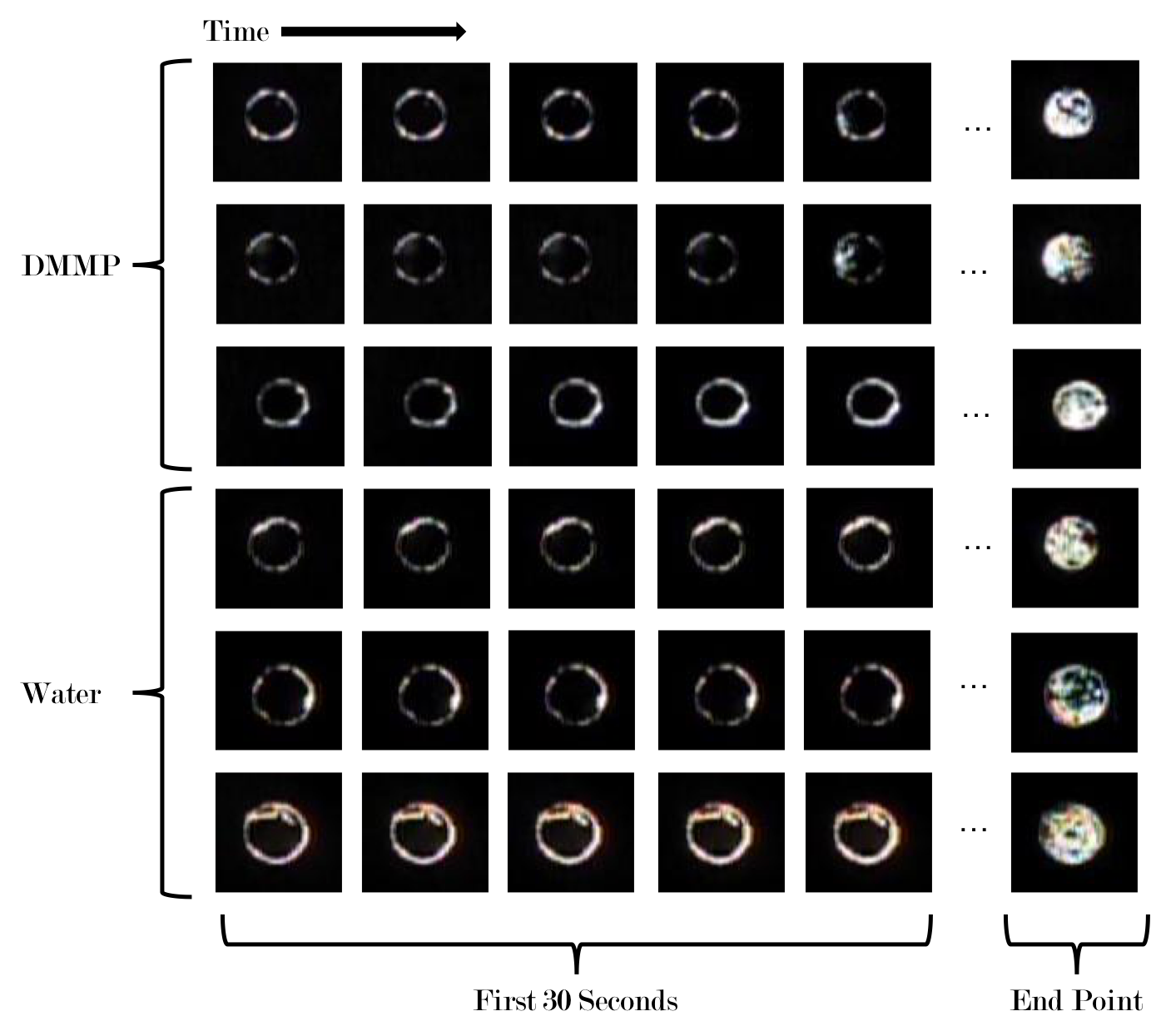} 
   \caption{Optical responses of liquid crystals under gaseous $\textrm{N}_2$-water (30\% relative humidity) and $\textrm{N}_2$-DMMP (10 PPM) environments. LCs were deposited into microwells with a diameter of 3mm to enable high-throughput data collection.  Reproduced from \cite{smith2020convolutional} with permission from the American Chemical Society.}
   \label{fig:timelapse}
\end{figure}

A challenge in the development of LC sensors is their sensitivity to {\em interfering} species. For instance, LC sensors designed for detection of dimethyl methylphosphonate (DMMP), \mynote{CH\textsubscript{3}PO(OCH\textsubscript{3})\textsubscript{2}}, might exhibit similar optical responses when exposed to humid nitrogen \cite{yang2005use}. These issues are illustrated in the experimental responses shown in Figure \ref{fig:timelapse}.  Our goal here is to determine whether or not one can unravel {\em hidden patterns} in the optical responses that can help discern between chemical species.  

We designed a CNN for automatically detect contaminant presence from optical LC images (micrographs). In summary, the approach us based on transfer learning; specifically, we used {\tt VGG16}, which is a pre-trained CNN for extracting features from the optical responses and these features were fed into a simple support vector machine (SVM) that predicts if the contaminant is present.  The results indicate that features extracted from the first and second convolutional layers of {\tt VGG16} allow for a {\em perfect} classification accuracy. The results also indicate that complex {\em spatial color patterns} are developed within seconds in the LC response, which leads us to hypothesize that fluctuations in LC tilt orientation (angle) play a key role in sensor selectivity.  Moreover, the analysis reveals that color is a key source of information that the CNN is looking for. 
\\

The dataset analyzed in this study was obtained from six videos that show the response of LCs to a gaseous stream of $\textrm{N}_2$ containing 10 ppm DMMP and six videos that show the response of LCs to a gaseous stream of $\textrm{N}_2$ containing 30\% relative humidity (both at room temperature). Each video tracks the dynamic evolution of multiple independent microwells (the total number of microwells recorded was 391). We split each frame into several images, each containing a single microwell at a specific time. The total number of microwell snapshots generated was 75,081. 

\begin{figure}[!htp] 
   \centering
   \includegraphics[width=0.7\textwidth]{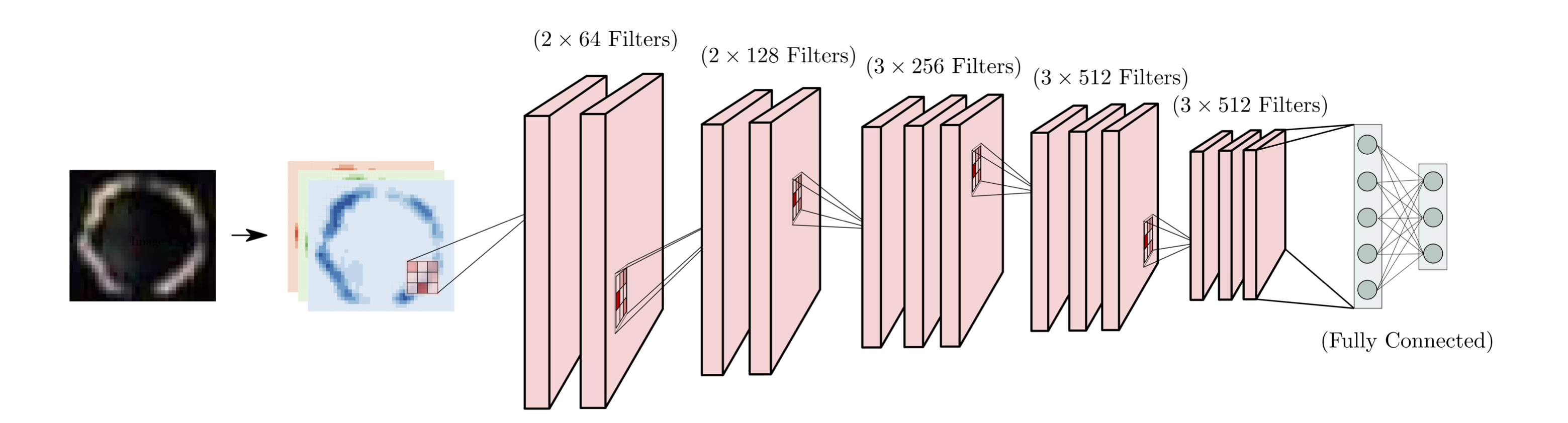} 
   \caption{\mynote{Schematic of pre-trained {\tt VGG16} architecture used for feature extraction of LC micrographs.} Reproduced from \cite{smith2020convolutional} with permission from the American Chemical Society.}
   \label{fig:VGG16}
\end{figure}

\begin{figure}[!htp] 
   \centering
   \includegraphics[width=0.7\textwidth]{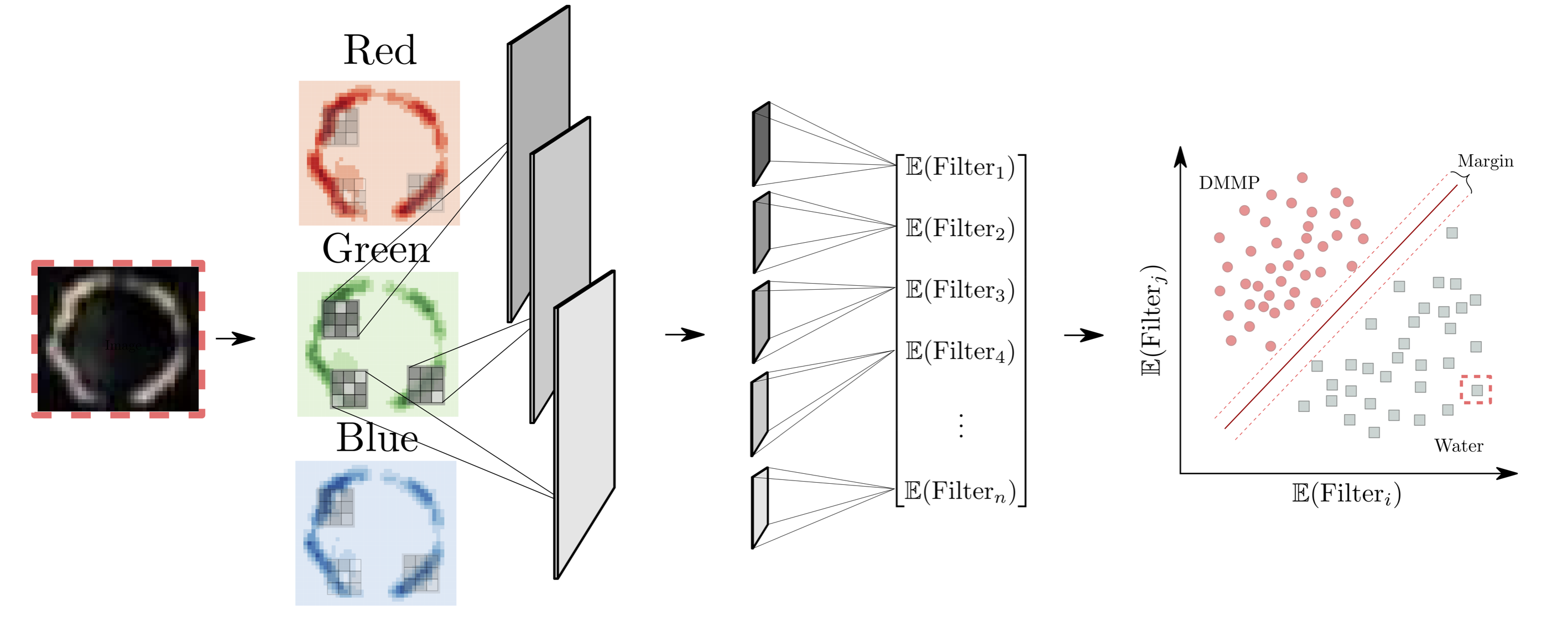} 
   \caption{\mynote{Schematic of ML framework including feature extraction and classification based on SVM.}  Reproduced from \cite{smith2020convolutional} with permission from the American Chemical Society.}
   \label{fig:feature_extr2}
\end{figure}

{The trained {\tt VGG16} network is freely available through the {\tt Keras} software and are what is used during feature extraction} \cite{chollet2015keras}.  This highlights the fact that highly sophisticated CNNs are openly available nowadays for conducting feature extraction.  A simplified representation of the {\tt VGG16} architecture is shown in Figure \ref{fig:VGG16}. The {\tt VGG16} network has been pre-trained to classify highly complex images from the internet (not related to our particular application, such as cat and dogs) and the deepest layers have been carefully tuned to differentiate such images. The early layers of the network are the most general and are easier to interpret; accordingly, we use the outputs of the first and second convolutional blocks to inform features for LSVM classification.  A visual representation of this process for the first  and second convolutional blocks is provided in Figure \ref{fig:feature_extr2}.  
\\

The framework using {\tt VGG16} features and LSVM was able to classify water and DMMP micrographs with {\em 100\% accuracy}. This result was achieved when using all of the 128 features of the second convolutional layer.  An accuracy of 98\% is obtained when we use the 64 features of the first convolutional layer. These results indicate that LC features that develop {\em early in the sensor response} are highly informative and sufficient to discriminate among chemical environments. 

\subsection{Molecule Design}

ML techniques have been recently used for property predictions of molecules such as water solubility \cite{huuskonen1998aqueous,lusci2013deep,boobier2020machine}, toxicity \cite{mayr2016deeptox,banerjee2018protox,jiang2021ggl}, and lipophilicity \cite{schroeter2007machine,tang2020self}. A fundamental step in the use of ML algorithms is the pre-definition or pre-calculation of molecular descriptors \cite{morgan1965generation,rogers2010extended,karelson1996quantum,moriwaki2018mordred}; such descriptors are used as input data to develop quantitative structure-property relationship models \cite{lo2018machine}. Among the various ML techniques, graph CNNs (GNNs) \cite{duvenaud2015convolutional,gilmer2017neural} have gained special popularity because they can directly incorporate molecular graph representations, thus retaining key structural information on molecules while avoiding the need to pre-calculate/pre-define molecular descriptors for which density functional theory (DFT) or molecular dynamics (MD) simulations may be required.  Overall, GNNs have shown strong predictive power in molecular property predictions, and have great potential to be applied to other fields for more accurate model development as well as enabling high-throughput screening of materials for manufacturing. 
\\

We show how to use GNNs for predicting critical micelle concentrations (CMCs) of surfactants. This study is based on the work presented in \cite{qin2021predicting}. When dissolved in water, surfactant monomers will undergo a cooperative aggregation process, called self-assembly, to form spherical micelles or related aggregate structures \cite{israelachvili2011intermolecular}. The formation of micelles in a solution can induce significant changes in key solution properties including the electrical conductivity, surface tension, light scattering, and reactivity \cite{rosen2012surfactants,israelachvili2011intermolecular}. Consequently, predicting conditions under which surfactants self-assemble is important for surfactant selection and design \cite{cheng2020design}. A critical parameter that characterizes surfactant self-assembly behavior is the CMC, which is the minimum surfactant concentration at which self-assembly occurs. 
\\

Conventionally, the CMCs are obtained experimentally by tensiometry, but it is laborious and expensive \cite{gaudin2016new,scholz2018determination,fluksman2019robust}. Here, we show that GNNs can predict CMC values directly from the molecular graph of a surfactant monomer. We gathered experimental CMC data measured at room temperature in water for 202 surfactants \cite{rosen2012surfactants,gaudin2016new,gahan2020bacterial,mukerjee1971critical} covering all surfactant classes. The proposed GNN architecture consists of two graph convolutional layers, one average pooling layer, two fully connected hidden layers, and one final output layer. A graph convolution layer updates each atom by aggregating the features of itself and its neighbors and maps the updated features into a hidden layer with 256 hidden features. The GCN model contains a total of 216,833 trainable parameters. 

\begin{figure}[!htp]
        \centering
        \includegraphics[width=0.6\textwidth]{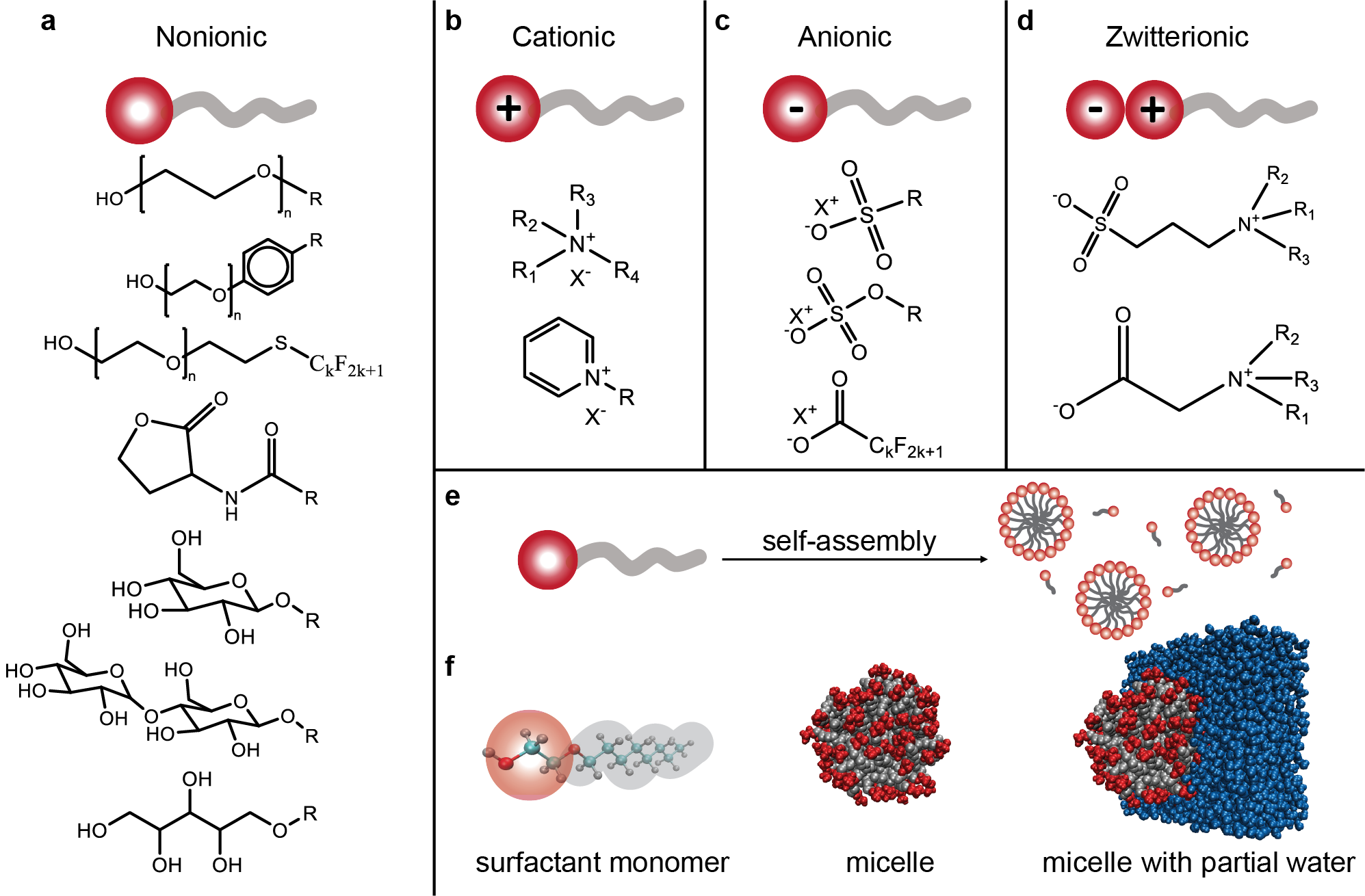}
        \caption{Overview of surfactant molecular structures and self-assembly process in micelles. (a–d) Sample structures of four classes of surfactants. Surfactants are categorized by the properties of their head groups as nonionic (a), cationic (b), anionic (c), or zwitterionic (d). (e) Surfactant monomers aggregate into spherical micelles in water with hydrophilic head groups facing toward the solvent and hydrophobic tail groups sequestered inside the micelle core. (f) Snapshots of a surfactant micelle from a representative MD simulation with water shown in blue. Reproduced from \cite{qin2021predicting} with permission from the American Chemical Society.}
        \label{surftype}
\end{figure} 

\begin{figure}[!htp]
        \centering
        \includegraphics[width=0.7\textwidth]{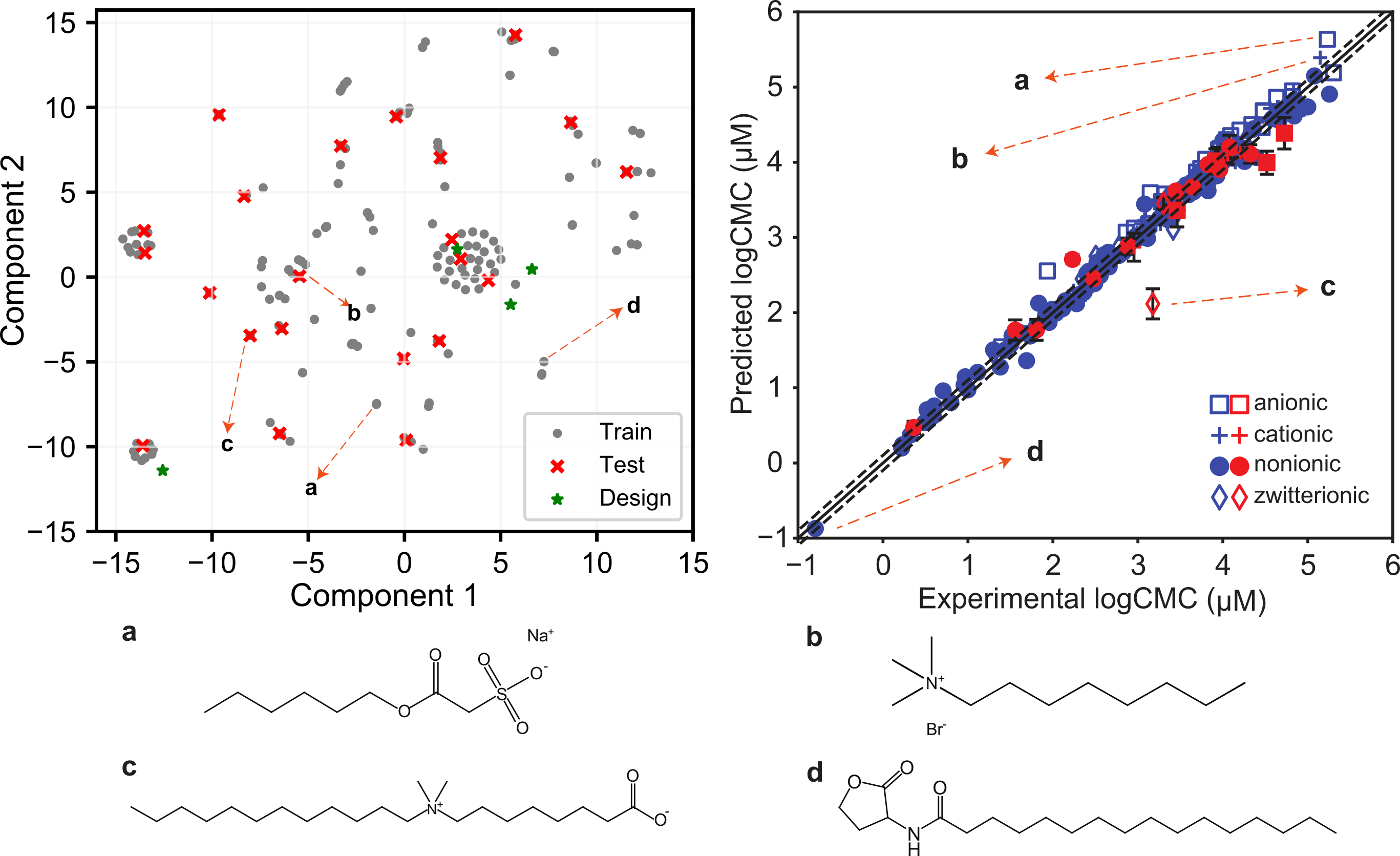}
        \caption{GNN predictions for all classes of surfactants. Left: low-dimensional distribution of surfactant fingerprints using t-SNE. The test samples (red crosses) are widespread, and most of the designed surfactants (green points) fall outside of the clusters of the existing data set. Right: parity plot between the predicted and experimental log CMC values (training data in blue and test data in red). The best-fit slope of the test data is 0.91 (R${}^2$ = 0.92), and the test RMSE is 0.30. Molecular structures are shown for the selected extreme points. Structure (a) is an anionic surfactant (minor outlier) with a high log CMC value. Structure (b) is a cationic surfactant (minor outlier) with a high log CMC value. Structure (c) is a zwitterionic surfactant (major outlier). Structure (d) is a nonionic surfactant with a low log CMC value. Reproduced from \cite{qin2021predicting} with permission from the American Chemical Society.}
        \label{cmcparity}
\end{figure} 

As illustrated in Figure ref{cmcparity}, the cross-validation (CV) RMSE on all classes of surfactants has a mean value of 0.39 with no significant outliers. We tested the model performance on a test data set, which contains samples from each surfactant class. We verified the distribution of the test samples using t-distributed stochastic neighbor embedding (t-SNE) \cite{van2008visualizing}, a nonlinear dimension reduction technique to visualize high-dimensional data, on the molecular fingerprints \cite{rogers2010extended} of the surfactants. Figure \ref{cmcparity} illustrates that the test samples are widespread, indicating the inclusion of unlike surfactant structures and classes in the test data set, which cover a much more diverse spectrum of surfactants than the data sets used in previous QSPR models.
\\

Figure \ref{cmcparity} also shows a parity plot between the experimental and predicted log CMC values for the training and testing sets. Cationic surfactants have the lowest test RMSE (0.07) followed by nonionic (0.18) and anionic (0.32) surfactants, and the model performs worst for zwitterionic surfactants (0.76). The parity plot also suggested a slightly lower accuracy for surfactants with relatively large log CMC values (>4.5). The overall predictability of the GCN model outperforms that of a prior QSPR model reported in the literature \cite{gaudin2016new}. The differences in the molecular structures found in our dataset highlights the wide variety of surfactants that the GCN model can capture.

\subsection{Decoding of Spectra}

We now show how to use CNNs to decode real-time spectra; specifically, we show how to characterize plastic components using real-time ATR-FTIR spectra. This case study also aims to illustrate how innovative (nonintuitive) data representations can be used to obtain more information from spectra.  This study is based on the work reported in \cite{jiang2021plasticnet}.
\\

Plastics are used in a wide range of applications such as food packaging, construction, transportation, health care, and electronics. Notably, only 20\% of all plastics produced were recycled \cite{owidplasticpollution}; this recycling rate is notably low compared to that of other materials (e.g., aluminum has a recycling rate of nearly 100\%). A key obstacle that arises here is the characterization of plastic components in mixed-platic-waste (MPW) streams.  ATR-FTIR (attenuated total reflection-Fourier transform infrared spectroscopy) is an instrumentation technique that can be used for analyzing plastic components found in MPW in real-time; as such, one can envision the development of fast, online ML techniques that can analyze ATR-FTIR spectra to characterize MPW streams. Here, we show that CNNs can characterize plastic components of MPW by decoding ATR-FTIR spectra. 

\begin{figure}[!htp] 
   \centering
   \includegraphics[width=0.5\textwidth]{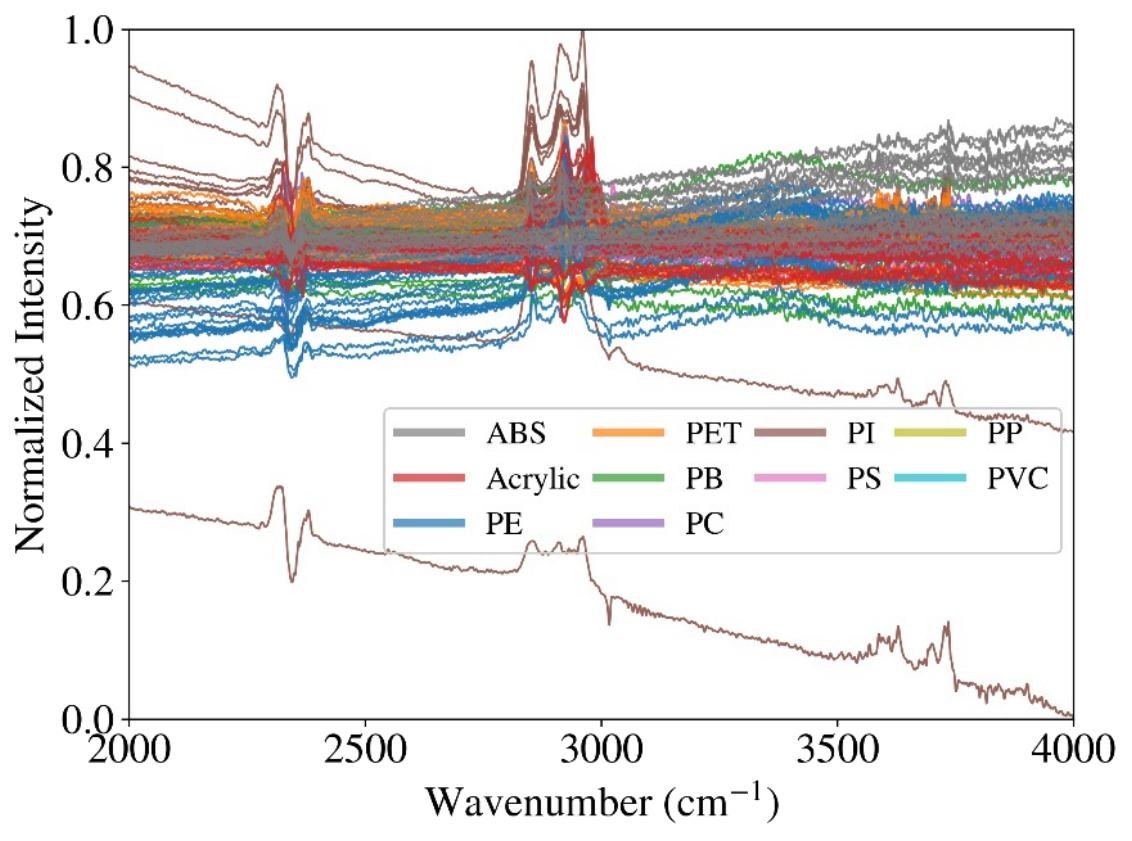} 
   \caption{Normalized infrared spectral intensities of various plastic materials. Each spectrum is a vector of length 4150. The resulting spectra contain significant noise and systematic errors. Reproduced from \cite{jiang2021plasticnet} with permission from Elsevier.}
   \label{fig:spectra_plasticnet}
\end{figure}

\begin{figure}[!htb]
	\begin{center}
		\includegraphics[width=0.6\linewidth]{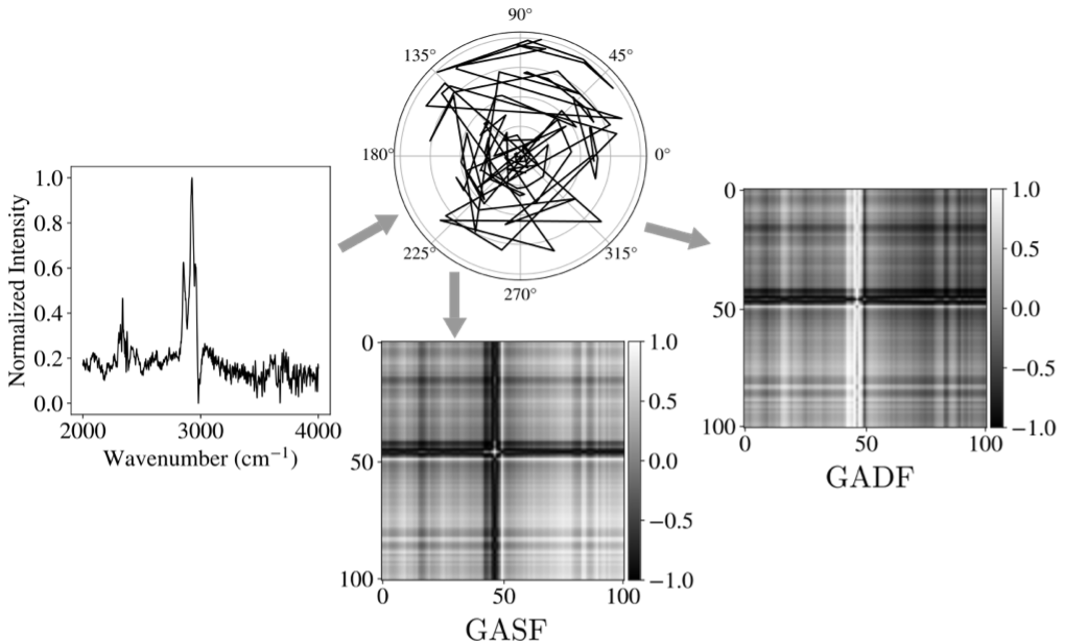}
		\caption{Conversion from 1D signal to GASF and GADF matrices. The 1D signal is first mapped to the polar coordinate system and finally converted to GASF and GADF matrices. Encoding the 1D signal into GAF matrices captures the relationship between the signal intensity at different wavenumbers. Reproduced from \cite{jiang2021plasticnet} with permission from Elsevier.}
		\label{fig:gaf_conversion}
	\end{center}
\end{figure}

Experimental data was obtained by preparing small sheets of plastics of different shapes and used ATR-FTIR to scan sheets for 10 different types; this data collection approach mimics how rigid waste plastics are found in online processing of MPW streams. The spectra collected can be represented as 1D vectors and analyzed by using 1D CNNs \cite{chen20191d}. The 1D CNN extracts features of a spectrum by convolving it with different filters; however, the 1D representation might fail to capture correlations across frequencies. Gramian Angular fields (GAF) have been recently used to encode 1D series data into matrices that capture correlation structures and that are processed using 2D CNNs; this data transformation approach has been shown to improve classification accuracy for time series data \cite{wang2015encoding}. A GAF represents vectors in a polar coordinate system and converts these angles into symmetric matrices using various operations. There are a couple of GAF types: Gramian Angular Summation fields (GASF) and Gramian Angular Difference fields (GADF). The conversion of spectra to GASF and GADF matrices is illustrated in Figure \ref{fig:gaf_conversion}; here, the GAF matrices are represented as grayscale images.

\begin{figure}[!htb]
	\begin{center}
		\includegraphics[width=0.6\linewidth]{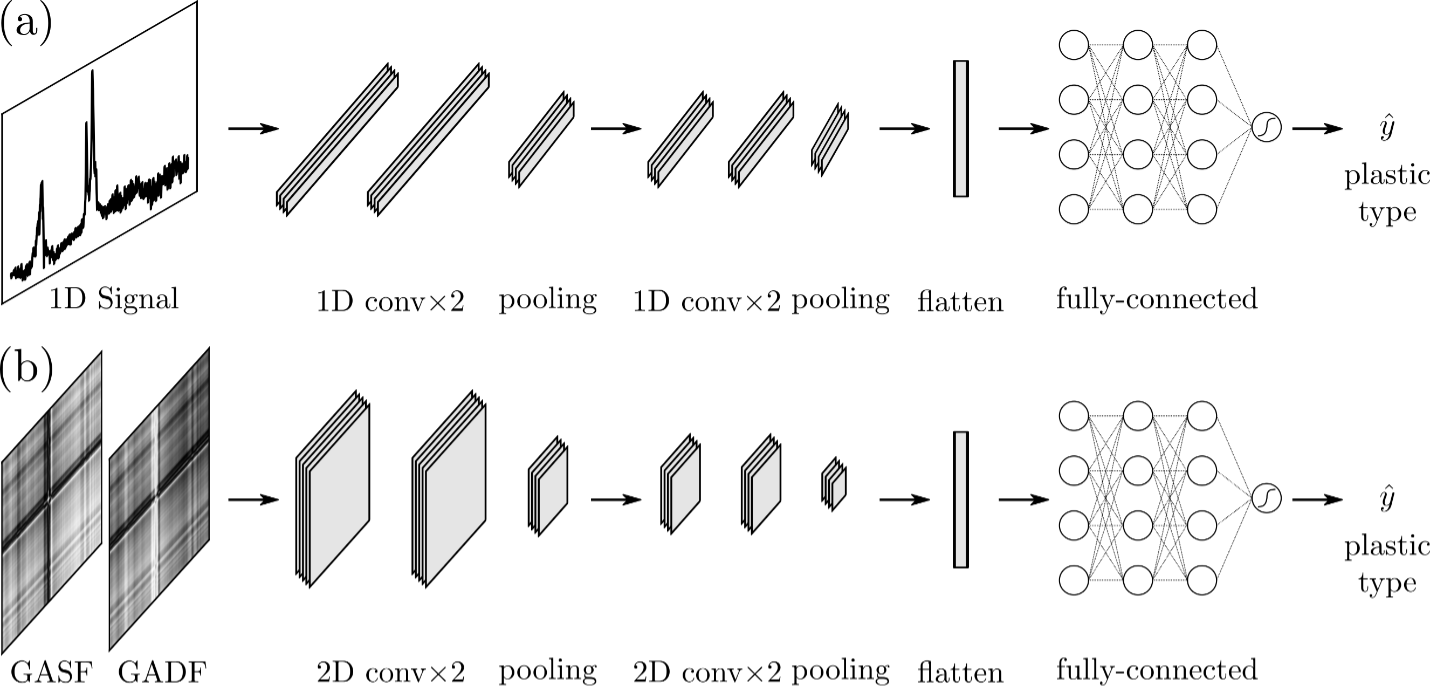}
		\caption{Architectures of (a) PlasticNet (1D) and (b) PlasticNet (2D). PlasticNet (1D) inputs a vector of 4150 and outputs the predicted plastic type. It contains 4 1D convolutional layers (each has 64 filters of dim 3), 1D max-pooling layers (each has a pooling window size of 2), a flatten layer, and 3 fully-connected layers (each has 64 units and a dropout ratio of 0.2). The activation functions between the layers are ReLUs. The final output activation function is softmax. PlasticNet (2D) inputs a GASF and a GADF matrix. The input size varies from $50 \times 50 \times 2$ to $250 \times 250 \times 2$. It has four 2D convolutional layers (each has 64 filters of $3 \times 3$), two 2D max-pooling layers (each has a pooling window size of $2 \times 2$). The flatten, fully-connected layers and activation function setups are the same as the ones of PlasticNet (1D). Reproduced from \cite{jiang2021plasticnet} with permission from Elsevier.}
		\label{fig:plasticnet}
	\end{center}
\end{figure}

\begin{figure}[!htb]
	\begin{center}
		\includegraphics[width=0.6\linewidth]{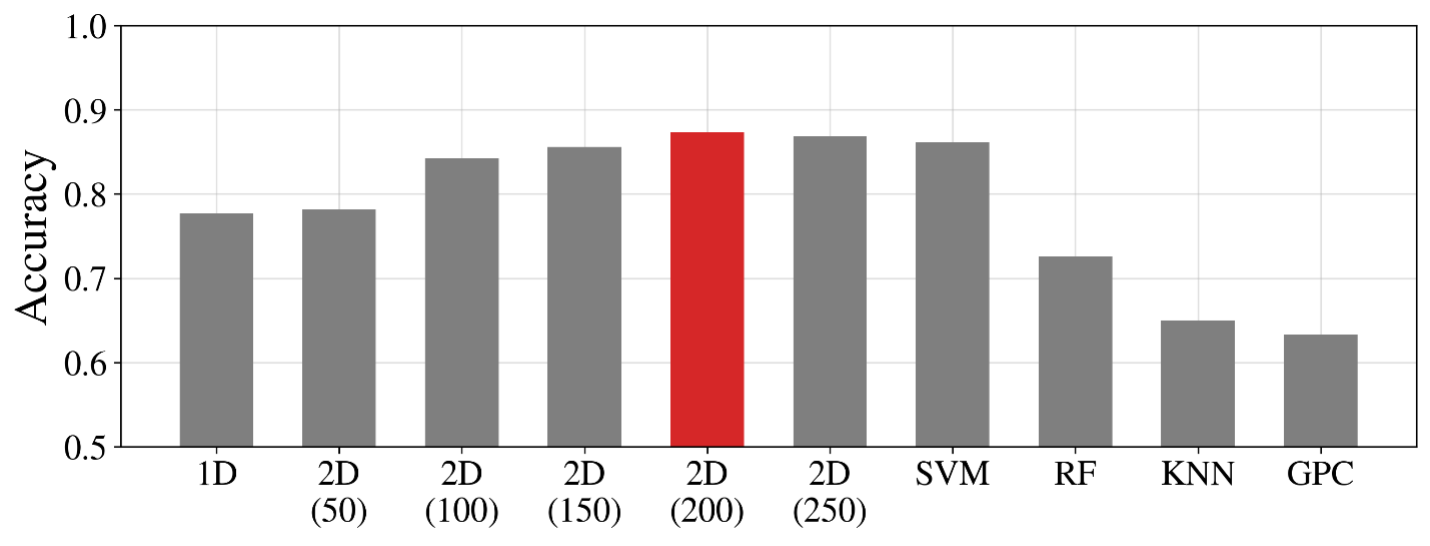}
		\caption{Comparison of the accuracy of CNN-based methods and other ML algorithms. PlasticNet (2D) with an input size of $200 \times 200 \times 2$ has the highest accuracy of 87.29\%. SVM with RBF kernels has a comparable accuracy of 86.14\%. The accuracy of PlasticNet (2D) is higher than that of PlasticNet (1D), indicating that the conversion from the original 1D signal to 2D GAF matrices captures more information. The accuracy of PlasticNet (2D) increases as the input matrix increases, indicating that a larger input matrix contains more information. Reproduced from \cite{jiang2021plasticnet} with permission from Elsevier.}
		\label{fig:plastic_result}
	\end{center}
\end{figure}

The comparison of architectures of PlasticNet (1D) and (2D) is shown in Figure \ref{fig:plasticnet}. Classification results for PlasticNet (1D) and (2D) are presented in Figure \ref{fig:plastic_result}. PlasticNet (2D) has a higher accuracy when the input size is larger than 100×100, compared to PlasticNet (1D) on raw IR spectra (77.7\%). Specifically, PlasticNet (2D) increases the accuracy of the PlasticNet (1D) by 12.4\%; this confirms that correlation information in spectra is important for classification. To validate the effectiveness of the proposed CNN models, we compared them with four commonly used ML classifiers, including Radial Basis Function (RBF) based Support Vector Machine (RBF-SVM), Random Forest (RF), k-Nearest Neighbors (kNN), Gaussian Process Classifier (GPC). The accuracy of PlasticNet (2D) is slightly higher (~1\%) than that of RBF-SVM when the input size is larger than $200 \times 200$. This indicates that RBF-SVM is comparable to CNN-based methods; however, SVMs offer limited flexibility to capture different representations for IR data. The accuracy of all methods saturates at 87\%, which suggests that the dataset itself contains significant errors that neither the CNN-based nor the SVM methods can explain. In subsequent work, we have shown that decoding fast MIR (mid-infrared spectroscopy) spectra using CNNs can achieve a nearly perfect plastic classification accuracy \cite{zinchik2021accurate}.

\begin{figure}[!htp] 
   \centering
   \includegraphics[width=0.5\textwidth]{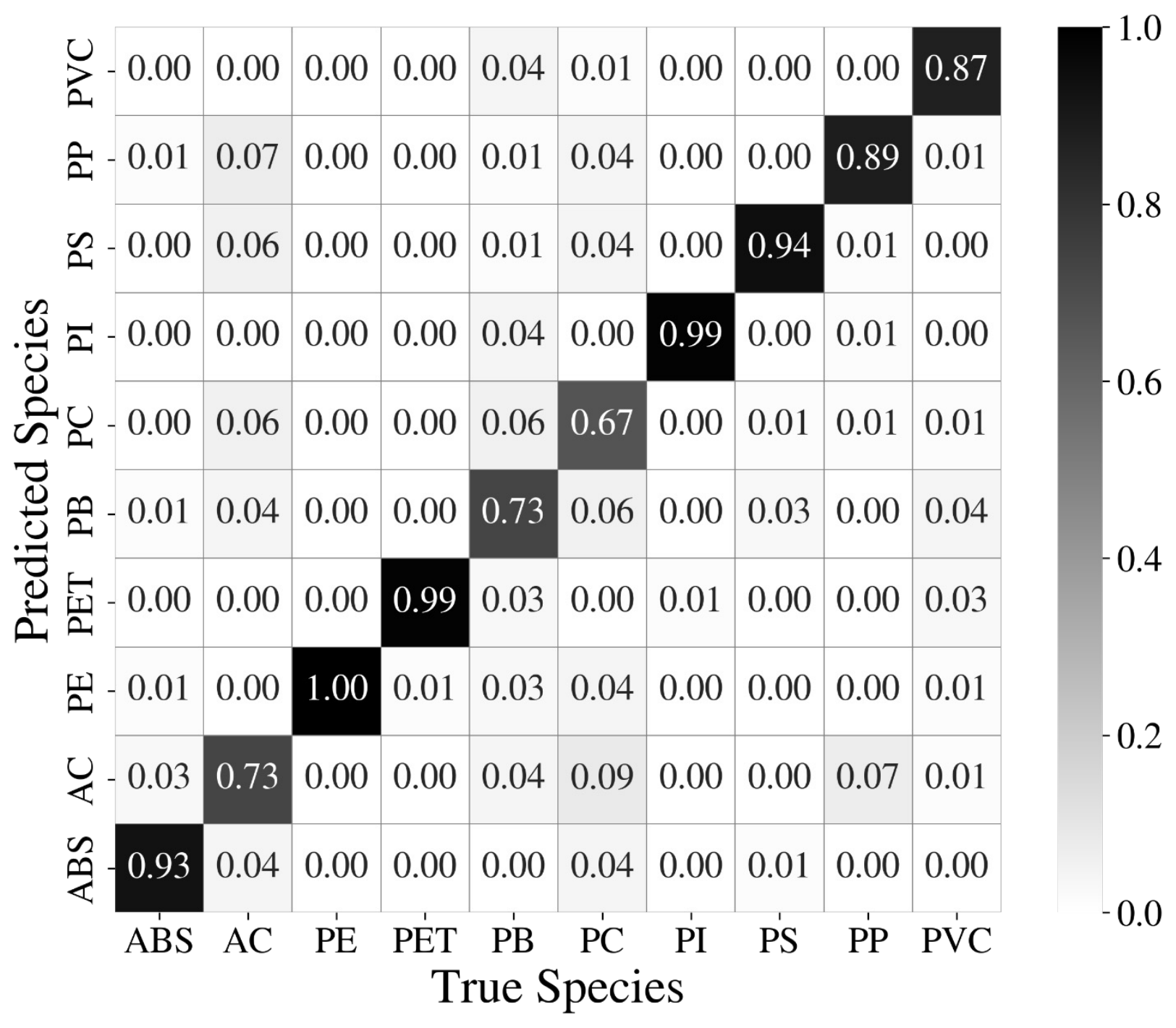} 
   \caption{Confusion matrix for PlasticNet (2D) with an input size 200×200×2. The overall accuracy is 87.3\%. Each column represents a true plastic species, and each row represents a model predicted plastic species. The entries along the diagonal are where the plastic species are correctly classified. Many diagonal entries are close to one, indicating that the PlasticNet (2D) has excellent classification accuracy. However, some plastic types cannot be classified with high accuracy (e.g., PC and AC). Reproduced from \cite{jiang2021plasticnet} with permission from Elsevier.}
   \label{fig:confusion_plasticnet}
\end{figure}

\subsection{CNNs for Multivariate Process Monitoring}

Multivariate process monitoring is a common task performed in manufacturing to identify abnormal/faulty behavior. Here, the idea is to collect multivariate time series (for different process variables) under different modes of operation (each mode is induced by a specific fault). The goal is to identify features (signatures) in the time series to determine if the process is a given mode. The case study presented here uses benchmark data for the Tennessee-Eastman (TE) process \cite{downs1993plant}. This study was based on the work presented in \cite{wu2018deep,jiang2021convolutional}. 

\begin{figure}[!htp]
	\begin{center}
	\includegraphics[width=0.7\linewidth]{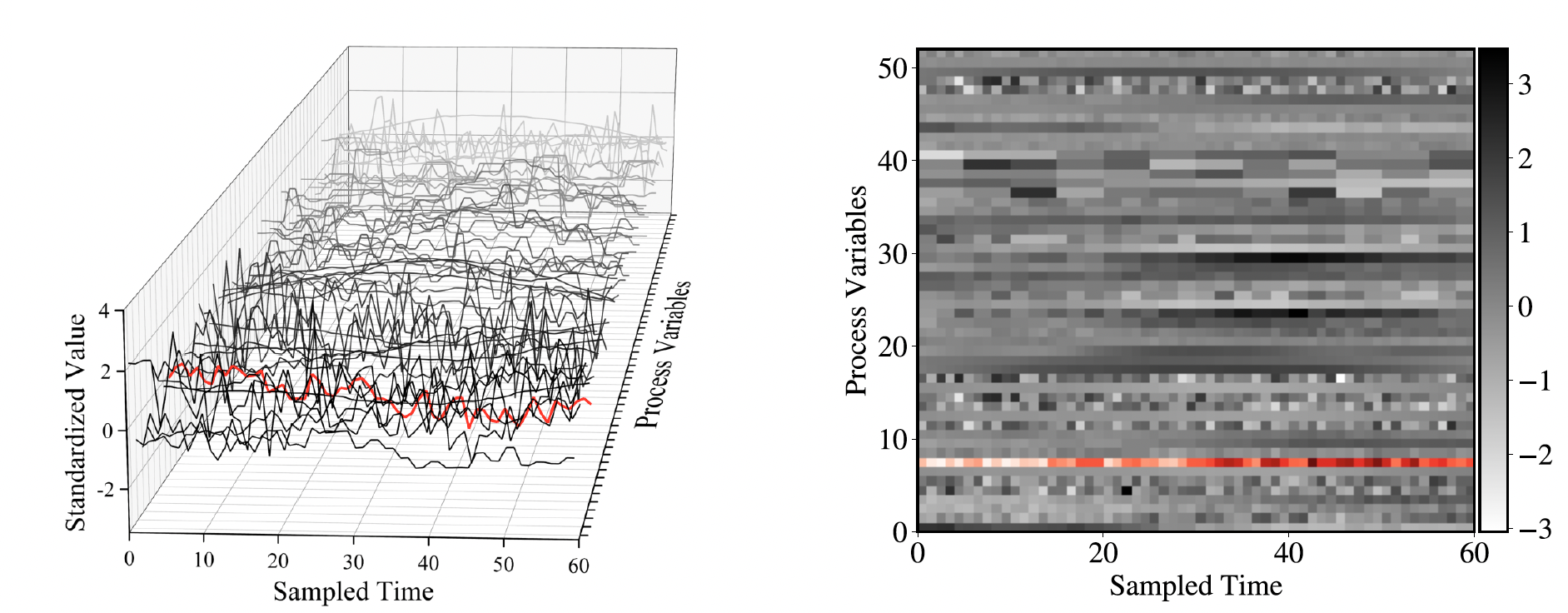}
		\caption{Representation of a multivariate time series as a 2D image. (a) 52 process variables collected 60 times over a 3-hour period. The variables are normalized with a zero mean and a unit variance. (b) An ${52 \times 60}$ matrix (visualized as an image). The red line in (a) and red row in (b) indicate the same data. Each row of the image represents one of the time series in (a). The fault number is 7 for (a) and (b), 9 for (c), and 15 for (d). We can see that (c) and (d) are visually similar but belong to different fault groups.  Reproduced from \cite{jiang2021convolutional} with permission from Wiley publishing.}
		\label{fig:case2signal2heatmap}
	\end{center}
\end{figure}

The TE process units include a reactor, condenser, compressor, separator and stripper. The TE process produces two products (G and H) and a byproduct (F) from four reactants (A, C, D and E). Component B is an inert compound. In total, the TE process contains a total of 52 measured variables; 41 of them are process variables and 11 are manipulated variables. This process exhibits 20 different types of faults related to changes in feed temperatures, compositions, reaction kinetics, and so on. 

\begin{figure}[!htp]
	\begin{center}
		\includegraphics[width=0.6\linewidth]{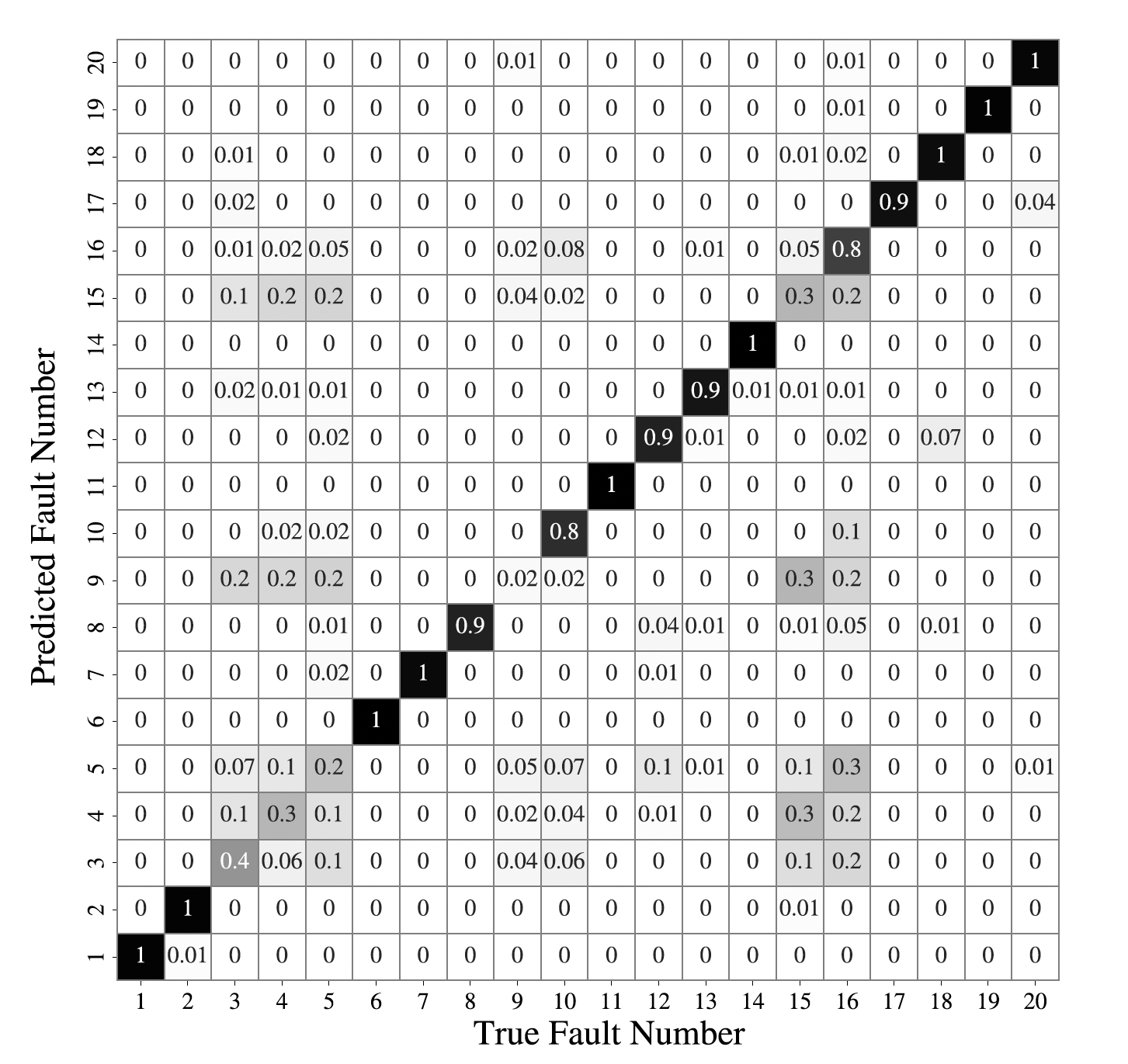}
		\caption{Confusion matrix from the CNN prediction. Each column represents a true fault type, and each row represents a CNN predicted fault type. The entries along the diagonal are where the fault types are correctly classified. Most of the diagonal entries are close to 1, indicating that the CNN has good classification accuracy. Reproduced from \cite{jiang2021convolutional} with permission from Wiley publishing.}
		\label{fig:case2confusion}
	\end{center}
\end{figure}

The TE process data is obtained from Harvard Dataverse \cite{DVN/6C3JR1_2017}. The 52 process variables are sampled every 3 minutes; the transformation of multivariate signal data to matrices is shown in Figure~\ref{fig:case2signal2heatmap}. We construct an input data sample by using 52 signal vectors (each vector contains 60 time points) that are combined into ${52 \times 60}$ matrix ($V$). We have a total of 6947 input samples and the training:validation:testing data ratio used is 11:4:5. Figure~\ref{fig:case2confusion} is the confusion matrix obtained for the CNN; the overall classification accuracy was 0.7561. With the exception of faults 3, 4, 5, 9, and 15, most faults can be identified accurately. Fault 3, 9 and 15 are especially difficult to detect because the mean, variance, and higher-order variances do not vary significantly.

\subsection{CNNs for Image-Based Feedback Control}

In this study, we consider techniques for incorporating CNN sensors (CNNs that map image signals to controllable state signals) in feedback control systems. We place a particular focus on the need for real-time novelty/anomaly detection approaches that provide robustness in effectively mitigating the consequences of visual disturbances. The concepts discussed here are a summary of the work presented in \cite{pulsipher2022safe}. This work in turn draws upon existing emergent applications of image data in control, as those presented in \cite{villalba2019deep, martynenko2017computer, rizkin2019artificial, lu2020image}. 
\\

\begin{figure}[!htb]
  \centering
  \includegraphics[width=0.7\textwidth]{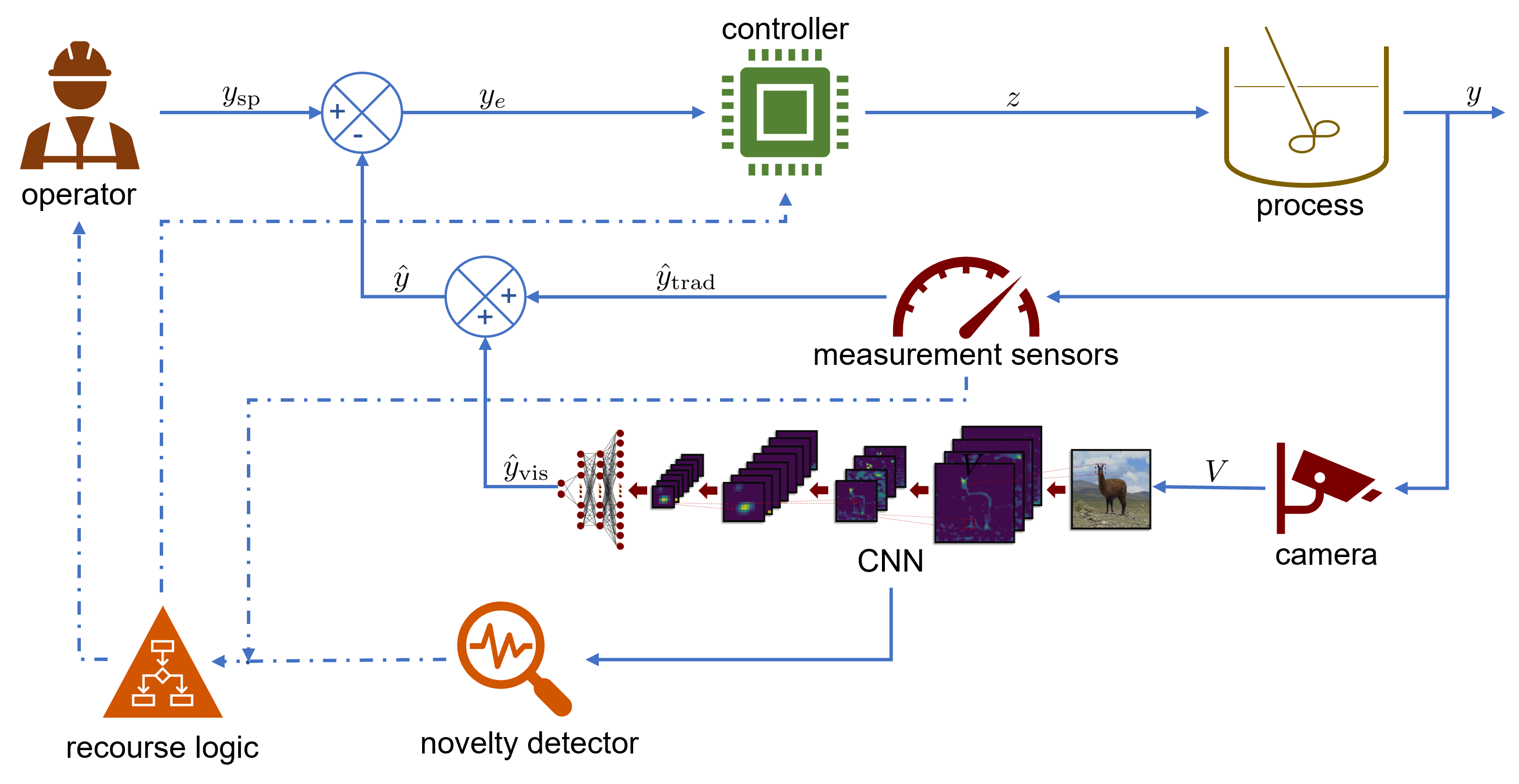}
  \caption{Feedback control system that incorporates a CNN sensor to convert image data into a controllable measurement signal that can be used for feedback control. Reproduced from \cite{pulsipher2022safe} with permission from Elsevier.}
  \label{fig:vision_control}
\end{figure}

\begin{figure}[!htb]
  \centering
  \includegraphics[width=0.7\textwidth]{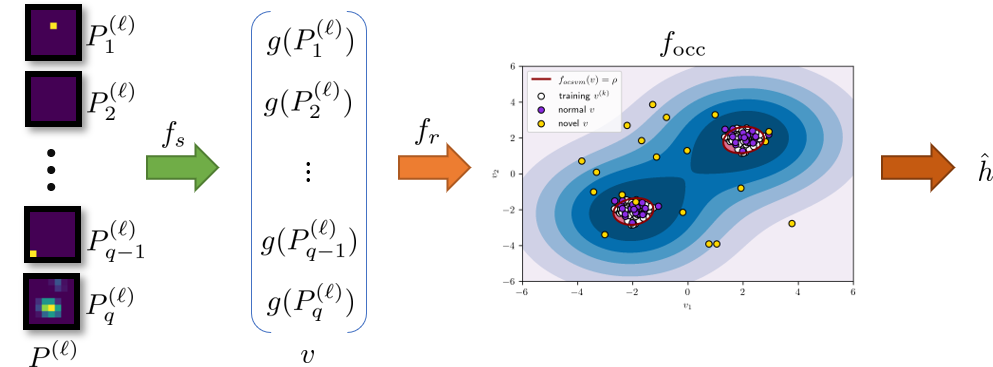}
  \caption{A summary of the SAFE-OCC novelty detection framework that operates on a feature map to produce a novelty signal. Reproduced from \cite{pulsipher2022safe} with permission from Elsevier.}
  \label{fig:feature_framework}
\end{figure}

We consider leveraging a CNN sensor to autonomously map image data to a controllable state variable such that we achieve closed-loop control. With this, we remove the human operator from the control loop in the sense that they will no longer need to actively interpret and act upon visual process data. Such a control system is depicted in Figure \ref{fig:vision_control}. Here, the camera and the CNN work together to form a computer vision sensor that is able to measure states $y_\text{vis}$ that otherwise would not be available using traditional process measurement devices. Hence, following this new paradigm we obtain a fully automatic control system that can exhibit improved setpoint tracking performance which promotes increased consistency and performance. 
\\

The paradigm shift from an operator-centric control system to the CNN-aided system of Figure \ref{fig:vision_control} introduces a significant vulnerability: poor prediction accuracy of $y_\text{vis}$ when the image $V$ is novel relative to the training data used to prepare the CNN (i.e., the CNN sensor makes a highly inaccurate prediction because it is extrapolating). Injecting erroneous measurement data into a feedback control architecture can have severe consequences. Thus, we require an appropriate novelty detection approach (depicted in Figure \ref{fig:vision_control}) to automatically recognize in real-time when the visual data $V$ is novel relative to the CNN sensor being used. Novelty detection denotes a set of unsupervised learning methods that differentiate between novel and normal data \cite{ruff2021unifying}. A couple of paradigms are reconstruction models and one-class classification.  One-Class Classification (OCC) denotes an area of methods that learn a single class of normal instances from unlabeled training data (typically assumed to be comprised of normal instances). These then identify novel data instances by determining if they lie outside the learned class. Here, we discuss the Sensor Activated Feature Extraction Once-Class Classification (SAFE-OCC) novelty detection framework \cite{pulsipher2022safe}. The SAFE-OCC framework leverages the native feature space of a CNN sensor to achieve novelty detection that is complimentary. The SAFE-OCC novelty detector involves three steps: feature extraction via the feature maps of a CNN sensor, feature refinement, and novelty detection via OCC (see Figure \ref{fig:feature_framework}). 

\begin{figure}[!htb]
   \centering
   \begin{subfigure}[b]{0.2\textwidth}
       \centering
       \includegraphics[width=\textwidth]{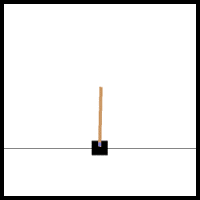}
       \caption{Simulation 1}
   \end{subfigure}
   \quad
   \begin{subfigure}[b]{0.2\textwidth}
       \centering
       \includegraphics[width=\textwidth]{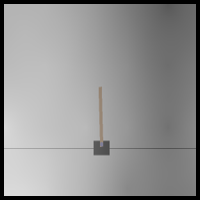}
       \caption{Simulation 2}
   \end{subfigure}
   \quad
   \begin{subfigure}[b]{0.2\textwidth}
       \centering
       \includegraphics[width=\textwidth]{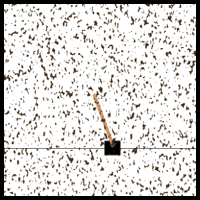}
       \caption{Simulation 3}
   \end{subfigure}
   \quad
   \begin{subfigure}[b]{0.2\textwidth}
       \centering
       \includegraphics[width=\textwidth]{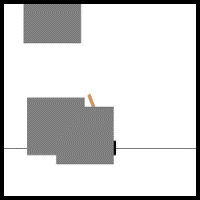}
       \caption{Simulation 4}
   \end{subfigure}
  \caption{Representative snapshots from simulations used in the cart-pole case study. Reproduced from \cite{pulsipher2022safe} with permission from Elsevier.}
  \label{fig:cart_images}
\end{figure}

\begin{figure}[!htb]
   \centering
  \begin{subfigure}[b]{0.45\textwidth}
      \centering
      \includegraphics[width=\textwidth]{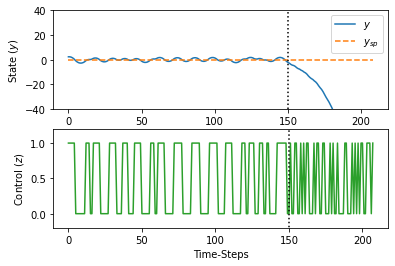}
      \caption{Simulation 4 Control Response}
      \label{fig:sim4_control}
  \end{subfigure}
   \quad
   \begin{subfigure}[b]{0.45\textwidth}
       \centering
       \includegraphics[width=\textwidth]{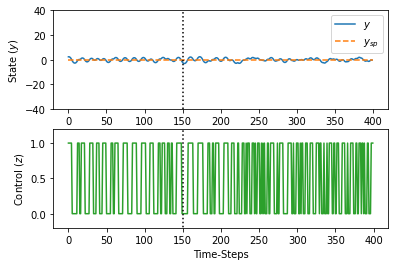}
       \caption{Simulation 2 Control Response}
       \label{fig:sim2_control}
   \end{subfigure}
   \\
 \begin{subfigure}[b]{0.45\textwidth}
      \centering
      \includegraphics[width=\textwidth]{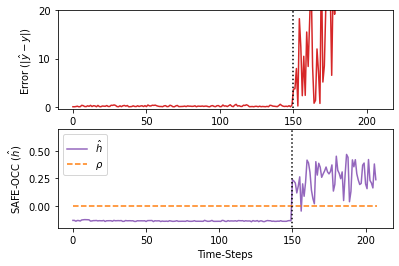}
      \caption{Simulation 4 Novelty Response}
      \label{fig:sim4_error}
  \end{subfigure}
   \quad
   \begin{subfigure}[b]{0.45\textwidth}
       \centering
       \includegraphics[width=\textwidth]{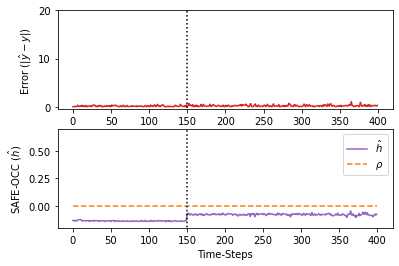}
       \caption{Simulation 2 Novelty Response}
       \label{fig:sim2_error}
   \end{subfigure}
  \caption{The control response trajectories of Simulations 2 (foggy disturbance) and 4 (blockage disturbance). The vertical dotted line at time-step 150 indicates when the fog disturbance is introduced in  Simulation 2 and the blockage disturbance is introduced in Simulation 4. Effective control is maintained in Simulation 2 because these are normal images for the CNN. Ineffective control, however, is obtained in Simulation 4 because the images are novel to the CNN. The SAFE-OCC framework detects this novel images to prevent this type of behavior. Reproduced from \cite{pulsipher2022safe} with permission from Elsevier.}
  \label{fig:cart_control_normal}
\end{figure}

We illustrate the application of SAFE-OCC to control the \texttt{CartPole-v1} environment from \texttt{OpenAI-Gym}\cite{1606.01540}, which corresponds to the classic cart-pole control problem introduced in \cite{barto1983neuronlike}. Here, we seek to balance a pendulum above a cart which we can move either right or left at a mixed rate. We consider the angle of the pendulum (measured in degrees relative to vertical alignment) as the state variable $y$ (ignoring the position of the cart), and we take the cart movement direction to be the control variable $z \in \{0, 1\} \subset \mathbb{Z}$ (where 0 is left and 1 is right). Thus, we have a single-input single-output (SISO) process to control. With this simplification, we implement a PID controller with a derivative filter. 
\\

We conducted four simulations: a base case that uses unperturbed images and three others that invoke a particular simulated visual disturbance after 150 time-steps. The three disturbance types are produced via \texttt{ImgAug} using the \texttt{Fog}, \texttt{Spatter}, and \texttt{Cutout} methods which correspond to fog, splattering, and square blockages, respectively. Figure \ref{fig:cart_images} shows representative images of these simulations. Figure \ref{fig:cart_control_normal} shows the responses exhibited in Simulations 4 (blockage) and 2 (fog). In Figures \ref{fig:sim2_control}, we observe that effective control in tracking the set-point is achieved for both simulations. This behavior can be attributed to the CNN sensor being trained with clear and fogged images which means that its predictions $\hat{y}$ incur a low error relative to $y$ as shown in Figure \ref{fig:sim2_error}. Figure \ref{fig:sim4_error} shows the response in the fourth simulation; here, the blockage disturbance is {\em novel} relative to the CNN sensor and thus, significant prediction error is incurred in each case once the sensor is subjected to the disturbance. SAFE-OCC accurately identifies the novel images once they are injected into the CNN sensor and can avoid catastrophic control failure. 

\section{Conclusion}

This chapter has reviewed the use of convolutional neural networks (CNNs) for extracting information from complex data sources that are commonly encountered in manufacturing.  We have used a set of selected case studies to demonstrate how these machine learning (ML) techniques can be used to conduct a variety of tasks such as classification, anomaly detection, and prediction of properties.   
\\

The discussion outlined in this chapter is heavily biased by research conducted by our research group and is not meant to provide an exhaustive review. The field of ML learning is quickly evolving, with many different applications and techniques being developed. For instance, transformer models are nowadays being developed for computer vision and provide an alternative to address limitations of CNNs \cite{carion2020end}. Moreover, CNNs are actively being used to tackle a wide range of challenges arising in catalysis, materials, healthcare, and biology \cite{haenssle2018man,bruno2018classification,cecen2018material}.  
\\

An area that is still in it infancy is the use of computer vision techniques to enable feedback control; specifically, closing the loop between image data and feedback control is a technical challenge. This is because computer vision signals are inherently infinite-dimensional objects (e.g., fields) that cannot be controlled directly by control systems. As such, it is necessary to extract key descriptors that effectively summarize these objects and it is necessary to develop techniques for building dynamical models directly from computer vision signals. Techniques such as autoencoders, recurrent neural networks, and dynamic mode decomposition provide some alternatives \cite{lu2020image,masti2021learning}. In this context, it is also important to re-think how to design control architectures that can properly act on image data, given that most manufacturing systems have limited actuation. This is particularly relevant in 3D-printing and additive manufacturing applications, in which it is necessary to achieve high level of precision in shaping 3D objects. Moreover, it is important to think about how to leverage these image data sources in the development of physical models \cite{li2018integration}.  The use of hyperspectral imaging in manufacturing is also an exciting direction \cite{botker2020hyperspectral}; this type of data can reveal properties of systems, materials, and products that provide rich information about quality and health. A fundamental challenge in dealing with hyperspectral imaging, however, is it inherit high dimensionality. Specifically, hyperspectral images contain many color channels and are challenging to process using CNNs. 
\\

Another exciting area of current research is speech recognition \cite{dimitrov2019autonomous}; one can envision a future in which instructions are provided to an automation system in the form of voice or in which the automation system summarizes the behavior of the system in narrative form or explains the reasoning behind a decision. Along these lines, the use of audio signals (e.g., as those obtained from vibration sensors) are currently being investigated to detect faults \cite{scheffel2021automated}.


\section*{Acknowledgments}

We acknowledge funding from the U.S. National Science Foundation under BIGDATA grant
IIS-1837812 and support from the members of the Texas-Wisconsin-California Control Consortium (TWCCC).


\bibliography{refs_bruce,refs_amy,refs_josh}

\end{document}